\documentclass{article}
\pdfoutput=1
\usepackage{TEMPCONF}
\usepackage{times}
\usepackage{epsfig}
\usepackage{graphicx}
\usepackage{amsmath}
\usepackage{amssymb}
\usepackage{units}
\usepackage{comment}
\usepackage{tabularx}
\usepackage{caption}
\usepackage{subcaption}
\usepackage{booktabs}
\usepackage{wrapfig}
\usepackage{siunitx}
\usepackage{authblk}

\usepackage[pagebackref=true,breaklinks=true,letterpaper=true,colorlinks,bookmarks=false]{hyperref}

\TEMPCONFfinalcopy 

\def\TEMPCONFPaperID{0000} 

\begin{document}
\pagenumbering{arabic} 
\title{On-manifold Adversarial Data Augmentation Improves Uncertainty Calibration}

\author[1,2]{Kanil Patel\thanks{firstname.lastname@de.bosch.com}}
\author[1]{William Beluch} 
\author[1]{Dan Zhang}
\author[1]{Michael Pfeiffer} 
\author[2]{Bin Yang} 

\affil[1]{Bosch Center for Artificial Intelligence, Renningen, Germany}
\affil[2]{Institute of Signal Processing and System Theory, University of Stuttgart, Stuttgart, Germany}

\maketitle
\begin{abstract}
Uncertainty estimates help to identify ambiguous, novel, or anomalous inputs, but the reliable quantification of uncertainty has proven to be challenging for modern deep networks. 
In order to improve uncertainty estimation, we propose On-Manifold Adversarial Data Augmentation or OMADA, which specifically attempts to generate the most challenging examples by following an on-manifold adversarial attack path in the latent space of an autoencoder-based generative model that closely approximates decision boundaries between two or more classes. 
On a variety of datasets as well as on multiple diverse network architectures, OMADA consistently yields more accurate and better calibrated classifiers than baseline models, and outperforms competing approaches such as Mixup, as well as achieving similar performance to (at times better than) post-processing calibration methods such as temperature scaling. 
Variants of OMADA can employ different sampling schemes for ambiguous on-manifold examples based on the entropy of their estimated soft labels, which exhibit specific strengths for generalization, calibration of predicted uncertainty, or detection of out-of-distribution inputs.
\end{abstract}

\section{Introduction}\label{sec:intro}
\begin{figure*}[!ht]
  \includegraphics[width=1.0\linewidth]{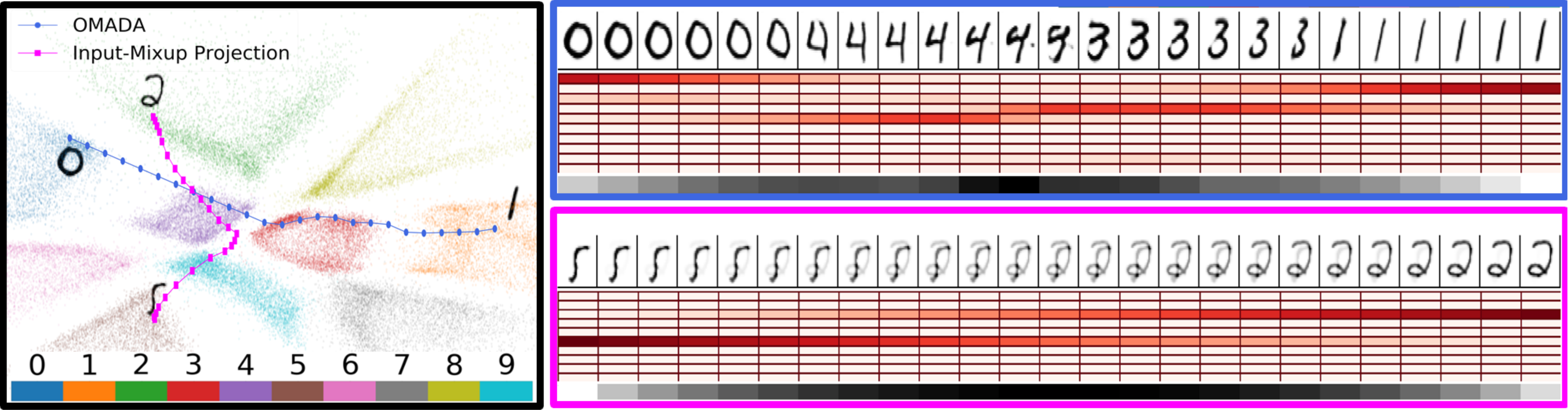}
  \caption{Visualization of an MNIST encoder-decoder latent space with two trajectories traversing between pairs of clusters. On the right we visualize the decoded image path for OMADA (top) and the Input-Mixup images (bottom) along with their corresponding soft labels (10 rows below images, red intensity corresponds to likelihood for classes $0$ to $9$), and the class entropy (bottom row, black shows high entropy).
    The OMADA trajectory starts at the cluster of ``0" and smoothly transitions to the target class ``1". It can be seen that the path favors routes which stick to the boundary regions of class clusters (e.g. going around the red cluster of ``3"s). Alternatively, we visualize the projection of Input-Mixup images onto the same manifold, for linear input interpolations between the digits ``5" and ``2".
  It can be seen that the images generated by OMADA are more confusing, and more importantly that the soft labels assigned to each image depend on the location on the manifold.
This is in contrast to Mixup, where the soft label will always be non-zero for all classes except the start and end interpolation points, regardless of whether the mixed images have similar features to other classes.  More OMADA trajectories can be found in the appendix.}
  \label{fig:teaser}
  \vspace{-0.575cm}
\end{figure*}

Deep neural networks (DNNs) have achieved spectacular success in classification tasks when trained on very large, but still finite training sets.
DNN training mostly follows the principle of Empirical Risk Minimization (ERM) \cite{vapnik_erm}, which states that by minimizing the training error the classifier will generalize to previously unseen data, under the condition that novel data points and labels are drawn from the same distribution as the training data.
Although this assumption works remarkably well on difficult benchmark datasets such as ImageNet \cite{russakovsky2015imagenet},
the assumption of identically distributed training and test sets is likely to be violated in DNN-based systems deployed in real-world situations.
Knowing when a DNN can or cannot be trusted because of dataset shift is of utmost importance whenever DNNs should be used in safety-critical applications \cite{Snoek2019,meinke2019towards}, such as autonomous driving, robotics, surveillance, or medical diagnosis.
At the same time, there can be true ambiguity in the data, e.g. when human annotators cannot agree or make mistakes \cite{Peterson:2019:HUM}, when inputs are corrupted or occluded, or whenever environmental conditions prevent a conclusive classification, e.g. due to challenging light or weather conditions \cite{hendrycks2019benchmarking}.
Such situations require DNNs that do not just predict the most likely class, but also quantify the uncertainty or confidence of their prediction, thereby allowing decision making systems to take the risk caused by perceptual uncertainty into account.

Unfortunately, the softmax outputs of modern DNNs, although accurate in their class predictions, have proven to perform poorly as indicators of uncertainty.
Overfitting to training data with one-hot encoded or \emph{hard} labels \cite{OnMixupTrainThul}, and over-confidence of ReLU networks for out-of-data inputs \cite{ReLUMHein} have been identified as potential root causes for this behavior.
Estimating the predictive uncertainty in deep learning is thus an active and challenging research topic.
The ultimate goal is to obtain \emph{calibrated} confidence scores \cite{GuoPSW17}, i.e. indicators that directly quantify the likelihood of a correct prediction.

\begin{figure*}
  \vspace{-0.80cm}
	\includegraphics[width=1.0\linewidth]{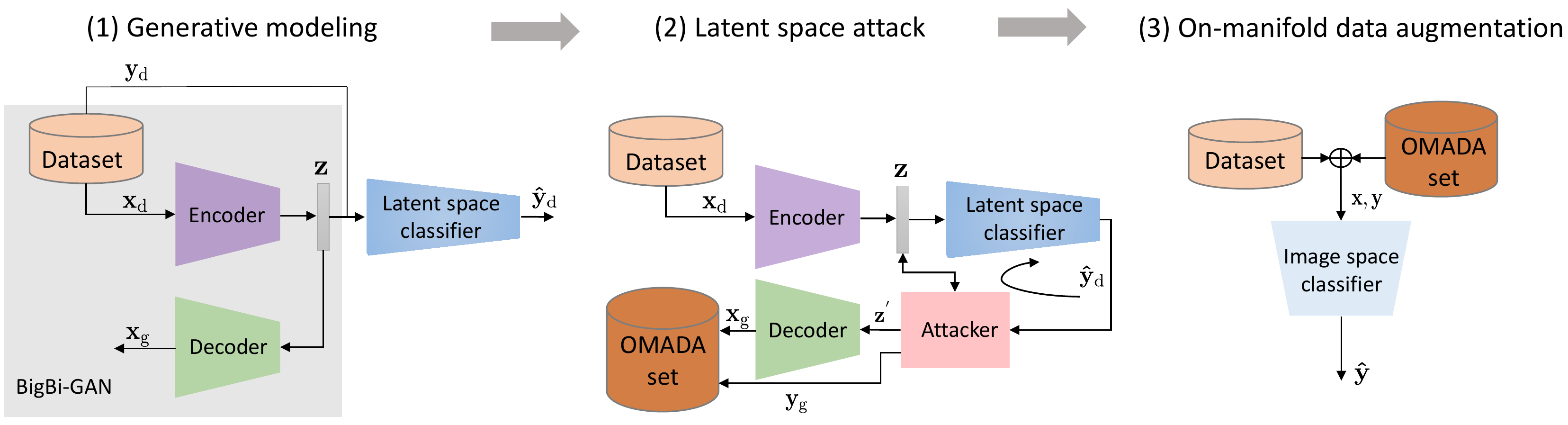}
	\caption{Illustration of the three phases of OMADA: (1) generative model, used in step (2) for latent space adversarial attacks to create the OMADA set, used in step (3) to train the classifier in image space using on-manifold data augmentation.}\label{fig:blockdiagram}
\end{figure*}

Since in all practical machine learning scenarios there is no access to the true data generating distribution, a reasonable starting point for uncertainty estimation is to assume that only data points in the vicinity of training data points can be predicted with high certainty \cite{Zhang2018mixupBE}. 
In fact, this is closely related to the problem of generalization, and various data augmentation techniques improve classification accuracy on unseen data by generating new training samples obtained by applying simple transformations to the original training samples without modifying the labels. 
In this article we propose a novel approach \emph{On-manifold Adversarial Data Augmentation}, or \emph{OMADA}, which yields calibrated uncertainty predictions by augmenting the training dataset with ambiguous samples generated by adversarial attacks \cite{goodfellow2014explaining}, but constrained to lie on an estimated training data manifold \cite{Stutz2019CVPR} (Fig.~\ref{fig:teaser}). 
The adversarial attack targets a latent space classifier. 
Unlike typical image-space classifiers that directly process the data samples, the latent space classifier is built on top of an autoencoder (encoder-decoder) based generative model (Fig.~\ref{fig:blockdiagram}), and processes the latent codes of data samples created by the encoder. 
OMADA can be viewed as a complementary approach to image-space attacks, which require the choice of an appropriate distance metric and $\epsilon$-ball in image space to keep the perturbed images realistic. OMADA instead considers only neighborhoods on the manifold by utilizing the generative model.

The encoder and decoder of the generative model are jointly trained to approximate the true data distribution.
By constraining the augmented samples to lie on the data manifold, we can closely approximate the true decision boundaries between classes by the latent-space classifier, while avoiding confusing the image-space classifier by injecting out-of-distribution samples into the training set.

We perform extensive experiments and comparisons against alternative methods from the literature for supervised classification. 
Augmenting the supervised classification training with OMADA, we observe significant improvements for calibration and accuracy across multiple benchmark datasets such as CIFAR-100, CIFAR-10 and SVHN and diverse network architectures such as DenseNet~\cite{dnpaper}, Wide ResNet~\cite{wrnpaper}, VGG~\cite{vggpaper}, and ResNeXt~\cite{resnextpaper}. 
Among all compared methods for uncertainty calibration, OMADA is the only one that consistently performs well across all benchmarks and architectures. The consistent failure on specific network architectures of simpler methods suggests that the quality of the generated confusing examples is extremely important to avoid inducing undesired artefacts during training.
Furthermore, we test the (image space) classifier on out-of-distribution samples. 
Using the confidence of the predictions as the metric to detect unseen data, OMADA outperforms multiple baseline methods in outlier detection performance in terms of the area under the ROC curve and Mean-Maximal Confidence (MMC). 
In summary, the results suggest that realistic but ambiguous on-manifold samples between one or more class clusters aid in resolving the notorious over-confidence characteristics associated with DNNs \cite{ReLUMHein}.

Our contributions include (1) a novel approach OMADA to create on-manifold ambiguous samples for data augmentation in supervised classification; 
(2) extensive empirical comparisons of a wide spectrum of alternative methods in the literature on various uncertainty evaluation metrics, and on out-of-distribution detection tasks; 
(3) extensive evaluation on a number of diverse network architectures;
(4) significant improvement over the benchmark methods on prediction calibration and outlier detection.  
For example, on CIFAR-100, OMADA results in up to a $9.8$x reduction in calibration error against standard training, up to a $5.9$x reduction compared to Mixup, and up to a $3$x reduction compared to temperature scaling.

\section{Related Work}~\label{sec:related work}
OMADA extends elements of recent successful approaches for uncertainty estimation, data augmentation, and adversarial training. 
The recently proposed \emph{Mixup} method \cite{Zhang2018mixupBE} creates new training samples by linear interpolation in the image space between a random pair of training samples.
In addition it also creates \emph{soft} labels by linearly interpolating between the original one-hot label vectors.
In \cite{OnMixupTrainThul} it was shown that Mixup not only improves generalization, but also yields well-calibrated softmax scores, and less confident predictions for out-of-distribution data.
A recent variant of Mixup is Manifold Mixup, which performs Mixup in the feature space of a DNN instead of the image space.
These methods may generate unrealistic samples that lie off the true data manifold.
Furthermore, the labels are generated by interpolating between two or more hard label vectors, and may thus not reflect true ambiguity, e.g. if the image obtained by interpolation is more similar to a third class (see Section~\ref{appendix_interpolation_methods} for an example).

Soft labels were also used to improve generalization via $\epsilon$-\emph{smoothing} \cite{epssmoothing}, where a probability mass of size $\epsilon$ is distributed over all but the correct class, thus penalizing over-confident predictions. 
Another simple and effective method to avoid over-confidence on outliers is to include out-of-distribution samples with uniform labels in the training set \cite{ReLUMHein,leeGANOutOfDistr}; the samples can even be as simple as uniform noise images.

Calibration can also be efficiently achieved after training, most prominently by \emph{Temperature Scaling} (TS) \cite{GuoPSW17}.
This method re-scales logits by a scalar chosen by minimizing the Negative Log-Likelihood loss on a validation set.
However, TS does not perform on par with other methods in outlier detection tasks, and when dataset shift occurs \cite{Snoek2019}. As a post-tuning method on a trained classifier, TS can be combined with data augmentation and label smoothing methods.

For studying adversarial robustness, the authors of~\cite{Stutz2019CVPR} introduced the concept of on- and off-manifold adversarial examples. Augmenting the training set with on-manifold adversarial examples is particularly useful to improve the generalization performance. However, common perturbations in the image space, including the above-mentioned Mixup, Manifold Mixup, and additive random noise, are not constrained to the data manifold. 
In this work we are interested in the use of on-manifold data augmentation for uncertainty calibration. 
The proposed OMADA method trains an autoencoder based generative model to approximate the data manifold and uses the adversarial attack in latent space to create ambiguous samples with soft labels. 
Unlike the soft labels created by $\epsilon$-smoothing, Mixup and its variants, the soft labels are semantically coherent with the samples, e.g., Fig.~\ref{fig:teaser}. 
In the experiments section, we will showcase the benefits of OMADA compared to these methods.

\section{OMADA Methodology}
The core of OMADA is constructing realistic, yet ambiguous samples for data augmentation, and input-dependent soft labels for improving the calibration of classifiers. 
This section explains in detail how to create on-manifold ambiguous training samples, and how to exploit them for the target classification task. 
Fig.~\ref{fig:blockdiagram} sketches the three main training phases of OMADA: generative modeling, latent space adversarial attacks, and classifier training with on-manifold data augmentation.

\subsection{Generative Modeling}\label{sec:methods_gen_model}
In order to model the complex high-dimensional space the data lies in, generative models are used to approximate the inaccessible ground truth data distribution.
We choose BigBi-GAN~\cite{bigbigan_donahue} because it has achieved state-of-the-art results on image synthesis and representation learning tasks, and exploit its design for learning the training data manifold. 

\paragraph{BigBi-GAN model}
The BigBi-GAN model (Fig.~\ref{fig:blockdiagram}-(1), ~\cite{bigbigan_donahue}) consists of an encoder $\mathcal{E}_{\rho}(z | x)$ and decoder $\mathcal{G}_{\phi}(x | z)$. The encoder encodes the input sample $x$ from the training set by a latent code $z$ that follows the standard normal distribution $P_{\mathrm{z}}$. The decoder attempts to reconstruct the input from the latent code $z$. 
The discriminator is trained to distinguish decoder outputs from real samples.
The decoder competes against the discriminator by synthesizing increasingly realistic samples. 
As the discriminator is only needed for training BigBi-GAN, it is omitted in Fig.~\ref{fig:blockdiagram}-(1).

\paragraph{Latent Space Classifier}
The current setting of the generative model is unsupervised. 
On top of it, we further introduce a latent space classifier $\mathcal{C}_{\gamma}(y | z)$ that exploits the label information to cluster the latent codes $\{z\}$ of $\{x_d\}$ according to the classes $\{y_d\}$ (Fig.~\ref{fig:teaser}). 
Namely, given the labeled training samples $(x_d,y_d)$, the classifier is trained by cross entropy minimization to predict the labels $y_d$ from the latent code $z_d = \mathcal{E}_{\rho}(x_d)$ obtained by applying the encoder to $x_d$. 
The cross entropy multi-class classification loss is added to the original encoder-decoder training loss of BigBi-GAN. 
The three networks are jointly trained to fool the discriminator. 
We further observe from Fig.~\ref{fig:teaser} that sampling from the boundaries between two class clusters yields ambiguous samples at the decoder output. 
Such generated samples lie in the support of the model distribution, which well approximates the data manifold. Therefore, they can be considered as on-manifold samples.

The trained encoder $\mathcal{E}_{\rho}(z | x)$, decoder $\mathcal{G}_{\phi}(x | z)$ and latent space classifier $\mathcal{C}_{\gamma}(y | z)$ provide all of the  necessary tools for OMADA to generate ambiguous samples and corresponding labels, which is described in the following section.

\subsection{Latent Space Adversarial Attack}
OMADA uses the generative model to synthesize samples which specifically have higher class ambiguity. 
Since ambiguous samples should reflect characteristics from two or more classes, their latent codes are expected to lie close to the class decision boundaries of the latent space classifier. 
As these latent codes of interest have an infinitely-small chance of being selected using conventional random sampling from the prior distribution on $z$, an alternative, novel sampling technique is required. 
Here we propose to use adversarial attacks on the latent space classifier to provide a targeted way to raise class ambiguity.

We start to explore the latent space from the latent code $z^s$ of an arbitrary training sample $x^s$ and move in a direction to approach a target class $y^o$. 
Here, $y^o$ is a one-hot vector encoding the class label. 
An adversarial attack, e.g. the projected gradient descent (PGD) method~\cite{pgdattack}, is used to find a small perturbation $z_{\mathrm{pert}}$ on $z^s$ such that the latent space classifier classifies $z^s+z_{\mathrm{pert}}$ as $y^o$ rather than $y^s$. Using the cross entropy loss, the perturbation $z_{\mathrm{pert}}$ is attained by solving the following minimization problem:
\begin{align}
z_{\mathrm{pert}}=\text{argmin}_{\Vert \delta\Vert_{\inf}} \sum_{i=1}^c (-y_{i}^o \text{log} \mathcal{C}_{\gamma}( z^s + \delta )_i),\label{eq:hardtgt}
\end{align}
where $c$ denotes the number of classes and $y_i^o$ is the $i$th entry of the one-hot vector $y^o$. Unlike standard adversarial attacks, here we do not need to constrain $\delta$ to lie within an $\epsilon$ ball. This is because the decoder is trained to produce realistic samples from any $z\sim P_{\mathrm{z}}$ and the support of the prior distribution $P_{\mathrm{z}}$ is the whole latent space. As depicted in Fig.~\ref{fig:blockdiagram}-(2), the work horse of our second phase training is the attack model to solve (\ref{eq:hardtgt}) in an iterative manner (for all adversarial attacks we perform $1$\unit{k} steps, using an $L_{\inf}$ norm with a step size $\alpha = 0.01$). The other networks in Fig.~\ref{fig:blockdiagram}-(2) are not changed after phase (1).

By iterating to solve (\ref{eq:hardtgt}), the intermediate results for $\delta$ added to $z^s$ create an attack path in the latent space (Fig.~\ref{fig:teaser}). Compared to simple linear interpolation in latent space, the proposed adversarial attack path has an important advantage: The adversarial loss (\ref{eq:hardtgt}) penalizes paths that pass through the class clusters except the target one. As shown in Fig.~\ref{fig:teaser}, the attacker mainly explores the empty regions between class clusters (i.e., decision boundaries of the latent space classifier) to reach the target, therefore being more efficient than linear interpolation in creating ambiguous samples. 
Feeding the latent codes along the attack path into the decoder $\mathcal{G}_{\phi}(z)$, Fig.~\ref{fig:teaser} depicts a series of synthetic samples that smoothly diverge from the source $x^s$ and approach a sample belonging to the target class $y^o$. The samples in-between realistically exhibit the features from both the source class $y^s$ and the target class $y^o$, and possibly other classes if they are encountered on the attack path. 

The labels of the samples can be obtained by applying the latent space classifier to the latent codes, i.e., $\mathcal{C}_{\gamma}( z^s + \delta)$. Unlike the one-hot encoded hard label vectors, the softmax responses can take on \emph{soft} values between $[0,1]$. Since the perturbation $\delta$ may traverse through multiple class boundaries to reach the target, the soft labels are not simply based on $y^s$ and $y^o$, and can have non-zero mass on other classes. Fig.~\ref{fig:teaser} shows that the soft labels are semantically coherent with the samples synthesized by the decoder. Comparing with Mixup~\cite{Zhang2018mixupBE}, which linearly interpolates both the samples and their labels, the proposed adversarial attack always produces on-manifold ambiguous samples and labels them according to the class-specific features.

Using the attacker together with the BigBi-GAN pretrained models to create our OMADA augmentation set, we investigate two ways to sample the latent codes from the attack path in the latent space. The first, and default mode, samples uniformly along the path. The second approach favors samples whose soft labels yield large entropies. After proper normalization, we use the entropies of each latent code's soft label vector along the path to parameterize a probability mass function (pmf), and then sample the latent code according to such a constructed pmf.

\subsection{On-Manifold Data Augmentation}
In order to solve the classification task, we train a DNN in the original input space $x$.
As shown in Fig.~\ref{fig:blockdiagram}-(3), the only difference is that we augment the original dataset with the OMADA set generated in Step (2) by sampling on data-manifold ambiguous samples together with their soft labels.
Combining the two datasets has two effects. Firstly, the enlarged training set improves the generalization performance and reduces model uncertainty. As the size of the OMADA set can be unlimited by repeatedly sampling the latent space, it also prevents overfitting and memorization. Secondly, the DNN learns from the soft labels of OMADA to make soft predictions in addition to hard ones, tempering overconfidence in the training process and achieving an improved calibration performance at test time. In the subsequent experiment section, we find that soft labeling of ambiguous samples is particularly helpful to detect out-of-distribution samples.

\section{Experiments}
\textbf{Setup} 
We evaluate and compare OMADA against multiple benchmark methods in the literature across $3$ datasets, i.e., CIFAR-100, CIFAR-10, SVHN and $4$ models, i.e., DenseNet ($L=100$, $k=12$) \cite{dnpaper},  Wide-ResNet 28-10 (WRN) \cite{wrnpaper}, ResNeXt-29 \cite{resnextpaper}, and VGG-16 \cite{vggpaper}. The benchmark methods from the literature primarily address data augmentation, label smoothing, and combinations of the two, similar to our proposed method. Additionally, we compare to Temperature Scaling (TS)~\cite{GuoPSW17}, as this is currently state-of-the-art for network calibration. 

The following is the list of methods we compare against: Base network (trained without data augmentation), Standard data augmentation (random crops and horizontal flips), Mixup ($\alpha=0.1$)~\cite{Zhang2018mixupBE}, Manifold Mixup ($\alpha=2.0$)~\cite{manifoldmixup}, $\epsilon$-smoothing ($\epsilon=0.1$)~\cite{epssmoothing}, CEDA~\cite{ReLUMHein} and TS. Unless otherwise noted, hyperparameters are taken from the original publications. For Mixup, $\alpha$ is chosen based on the results from ~\cite{OnMixupTrainThul}. Further details about hyperparameters for individual methods can be found in Section ~\ref{experiment_hyperparameters}.

\textbf{Training details} The training hyperparameters (learning rate, etc.) for each network are listed in the appendix (Section ~\ref{experiment_hyperparameters}); these hyperparameters do not vary across datasets and methods.  At the end of training, the model weights used for evaluation are chosen from the epoch with the best validation accuracy. 
Each reported result is the mean over $5$ independent runs with the same hyperparameters.

For all OMADA-trained networks we evenly balance each batch with $50\%$ of the real training samples and $50\%$ of the on-manifold adversarial samples. In order for these networks to be comparable to other baselines, we ensure that each epoch has the same number of updates as the Base method (i.e. for each epoch the OMADA-trained networks only observe $50\%$ of all real training samples).

\subsection{Tasks}
While the experimental investigation is primarily focused on calibration, we also look at other applications of network uncertainty, and the classification accuracy. 

\textbf{Calibration}
A classifier is well-calibrated if its probabilistic output corresponds to the actual likelihood of being correct, i.e. of all images a network predicts with a softmax confidence of $0.9$, approximately $90\%$ should be classified correctly.  This is typically measured by creating a Reliability Diagram, in which images are binned by the softmax value of their predicted class, and calculating some distance metric between the resulting curve and the ideal calibration curve.  The most popular of these metrics is the Expected Calibration Error (ECE) ~\cite{GuoPSW17}.  We instead use the Adaptive Calibration Error (ACE), which results in an equal number of images per bin; this metric is more robust wrt. binning hyperparameters and the baseline network accuracy ~\cite{ace_paper}: 

\begin{equation}
\mathrm{ACE} = \frac{1}{R}\sum_{r=1}^{R} | \mathrm{acc}(r) - \mathrm{conf}(r)| ,
\end{equation}
in which $N$ is the total number of data points, and the calibration range $r$ is defined by the $[N/R]^{\mathrm{th}}$ index of the sorted softmax predictions. 

\textbf{Outlier Detection} 
Outlier detection focuses on identifying out-of-distribution (OOD) inputs at test time, based on thresholding the predicted uncertainty. OOD data can be a completely different dataset, corrupted data, or classes from the same dataset not seen during training.  The outlier detection experiments in this paper focus on the former case; the network is trained on CIFAR-10, and the predicted softmax is used to try and identify anomalous SVHN images at inference time. The metric used for evaluating outlier detection performance is the area under the receiver operating characteristic (ROC) curve (AUC). Intuitively, this measures the ability of the uncertainty measure to binary classify an input as in-distribution or out-of-distribution over various thresholds.

\textbf{Other Uncertainty Measures} 
We also investigate how confident the networks are on OOD data by measuring the Mean Maximal Confidence (MMC), and how well the produced uncertainty estimates correlate with the true error by producing Sparsification plots~\cite{geifman2017coverage}. These results, and a more detailed explanation of the metrics, can be found in the appendix (Section~\ref{appendix_results}). 

\begin{figure*} [!ht]
	\centering
	\begin{subfigure}{.48\textwidth}
		\centering
		\includegraphics[width=0.95\textwidth, clip]{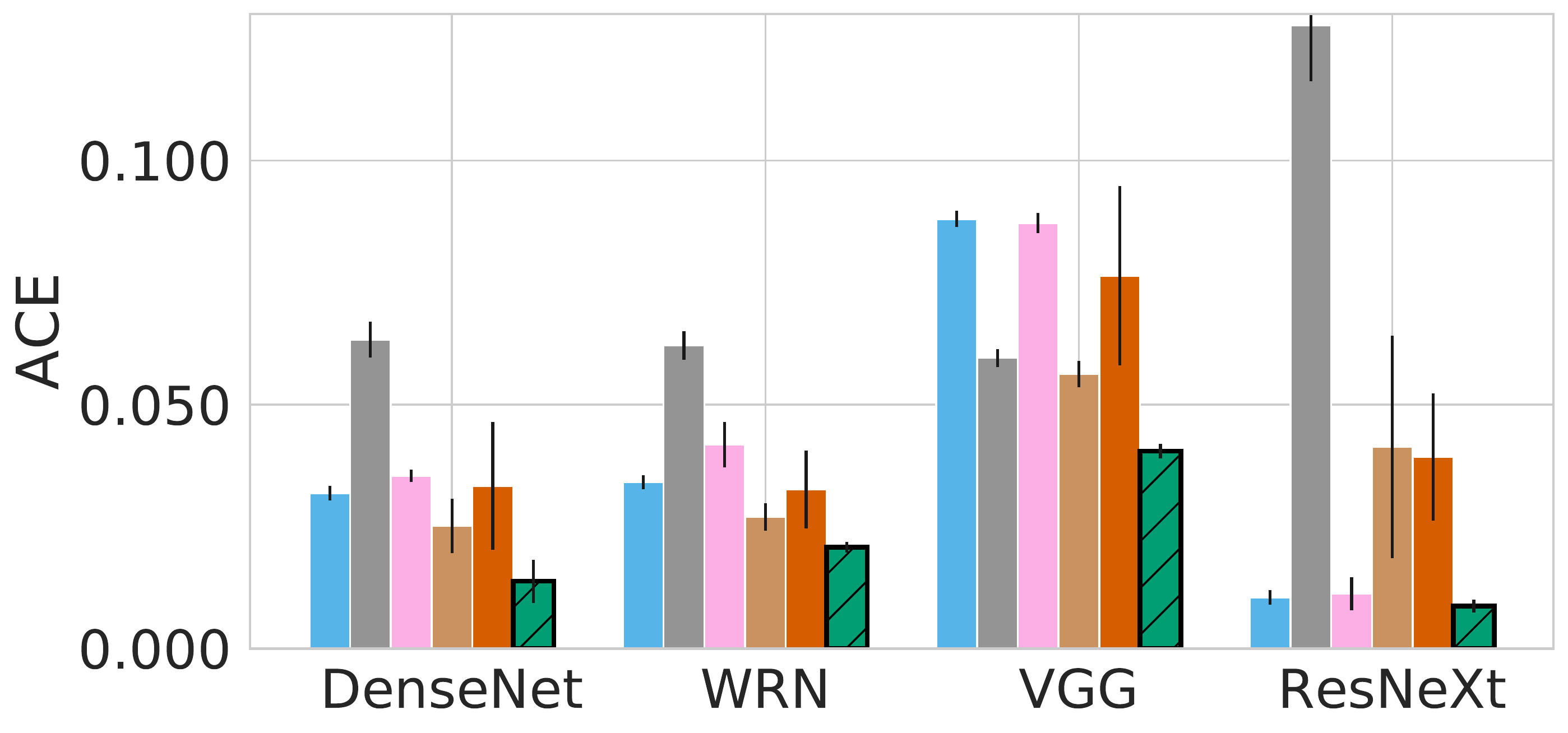}  
		\caption{CIFAR-10 ACE}
		\label{fig3a_cifar10_ace}
	\end{subfigure}
	\begin{subfigure}{.48\textwidth}
		\centering
		\includegraphics[width=0.95\textwidth, clip]{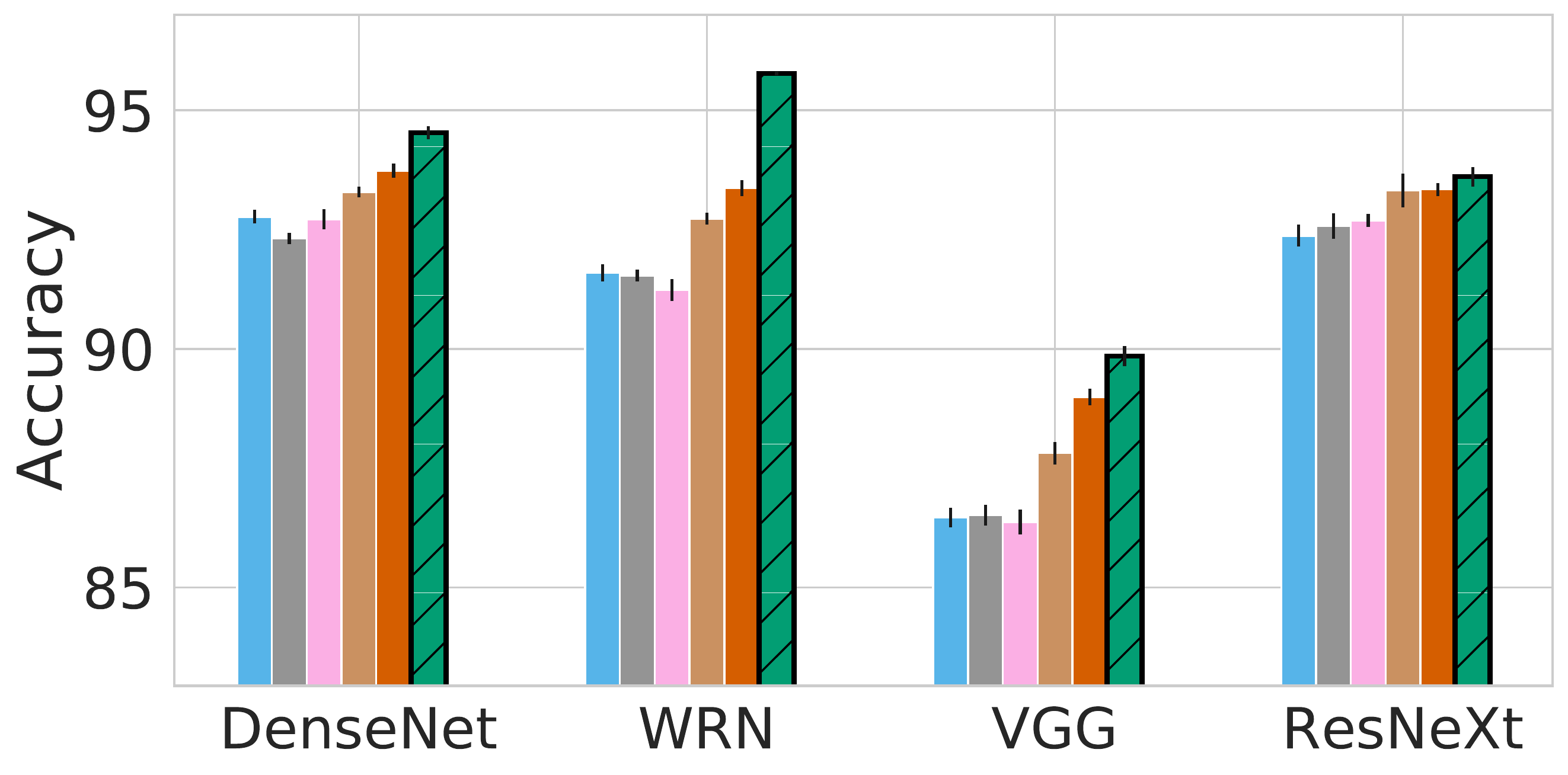}
		\caption{CIFAR-10 Accuracy}
		\label{fig3b_cifar10_acc}
	\end{subfigure}
	\begin{subfigure}{.48\textwidth}
		\centering
		\includegraphics[width=0.95\textwidth, clip]{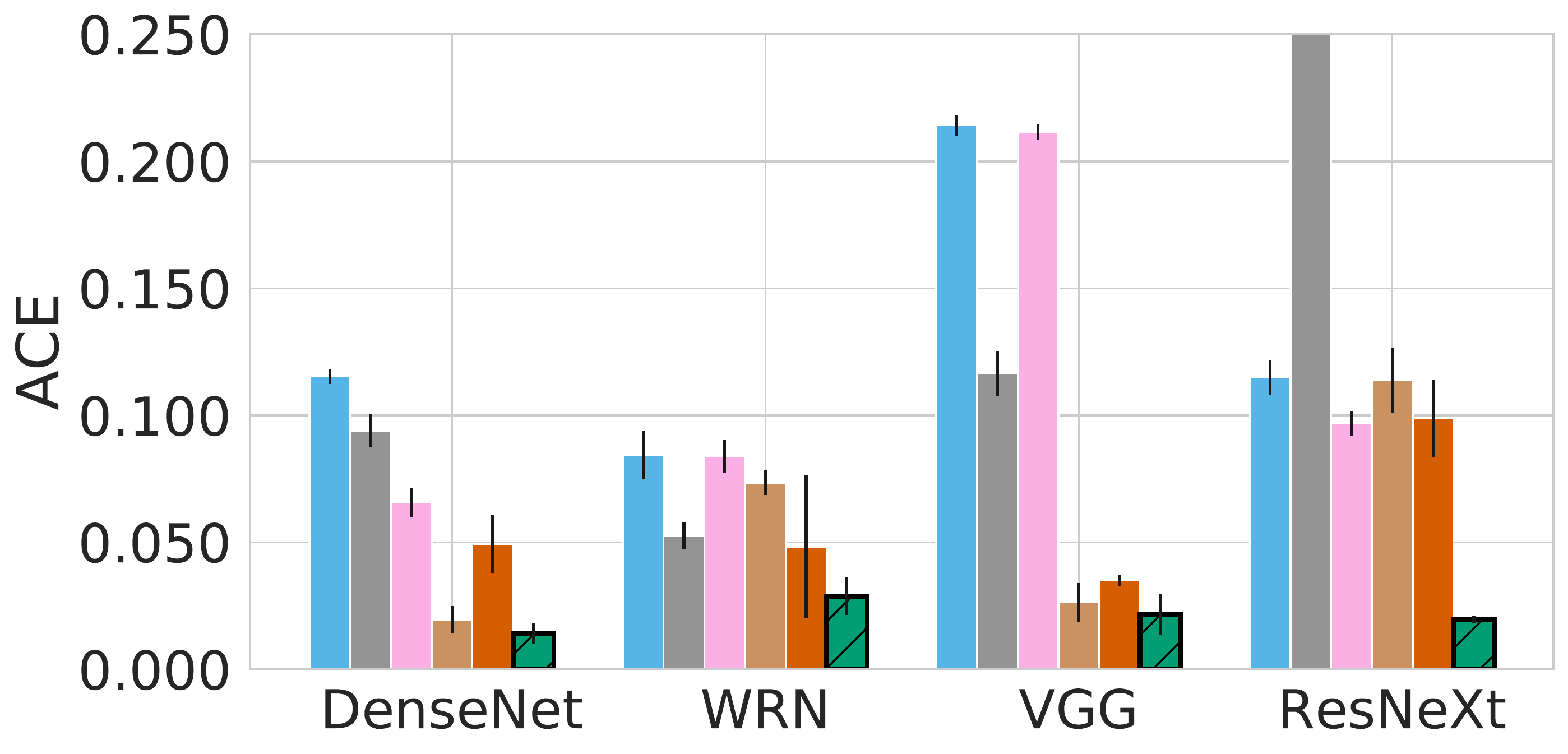}  
		\caption{CIFAR-100 ACE}
		\label{fig3c_cifar100}
	\end{subfigure}
	\begin{subfigure}{.48\textwidth}
		\centering
		\includegraphics[width=0.95\textwidth, clip]{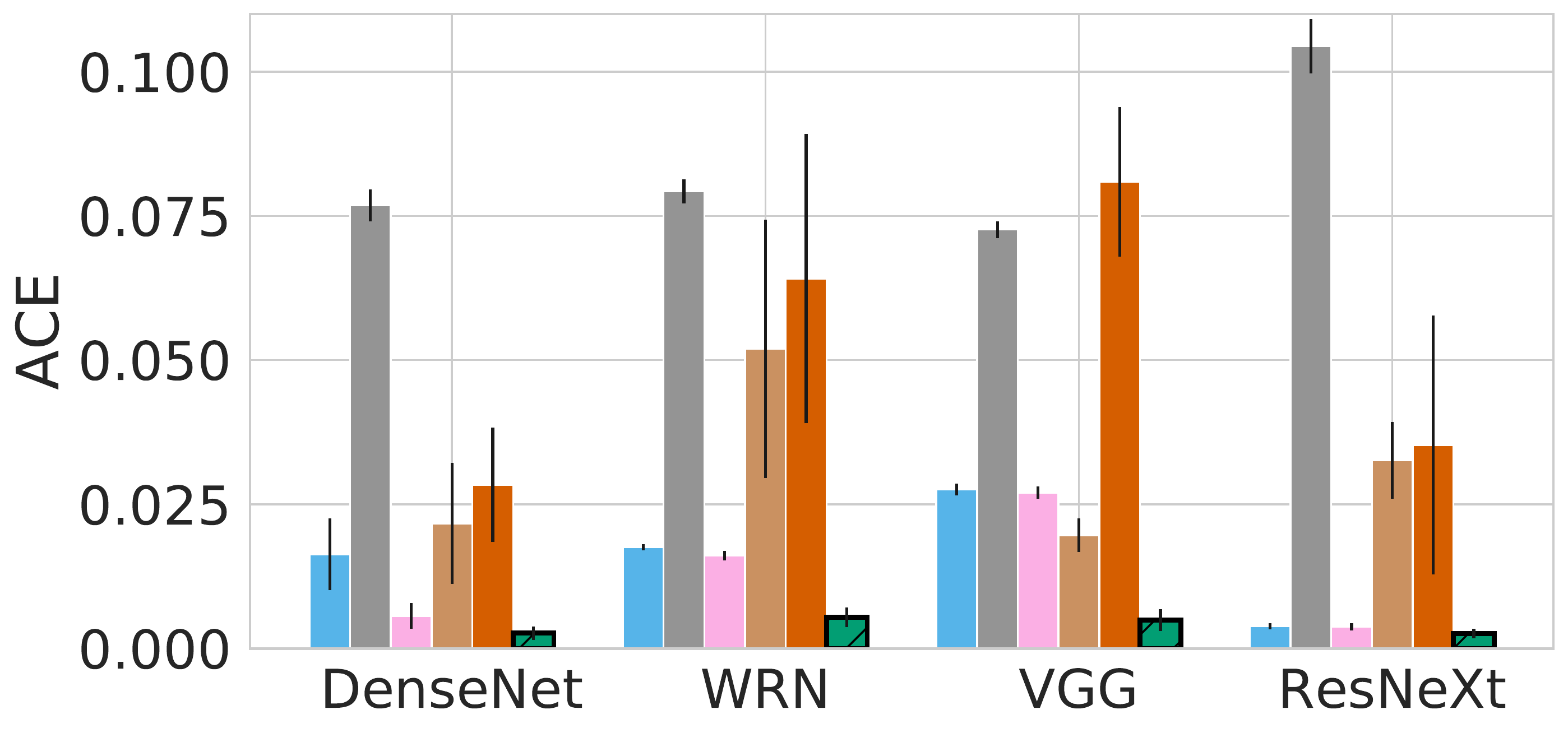}
		\caption{SVHN ACE}
		\label{fig3d_svhn} 
	\end{subfigure}
	
	\begin{subfigure}{.99\textwidth}
		\centering
		\includegraphics[width=0.95\textwidth, clip]{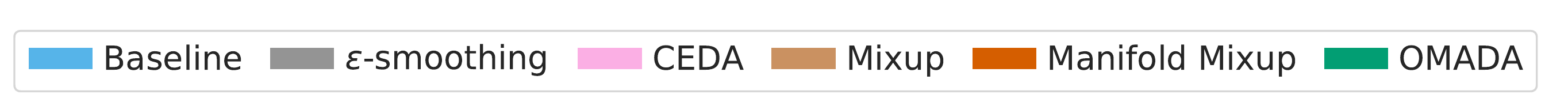}
		\label{legend}
	\end{subfigure}
	\caption{Calibration performance (ACE) of label-smoothing methods on in-distribution data for CIFAR-10 (a), CIFAR-100 (c), and SVHN (d).  Hatched bars indicate the best-performing method per network architecture.  Error bars are $\pm$ 1 std. dev. over 5 runs. (b) Shows the classification accuracy on CIFAR-10. Across all datasets and architectures, OMADA achieves better ACE, and higher accuracy than all other calibration methods.}
	\label{fig3_ace}
	\vspace{-0.5cm}
\end{figure*}

\subsection{Experimental Results}
\textbf{Network Calibration}
We first compare the calibration and accuracy performance of OMADA against a range of baselines and competing methods by using training with no standard augmentation as the baseline for all trained networks.
The reason for this choice is to have a controlled setting where each method's effect on both calibration and generalization can be isolated.
In order to be comparable against the literature, which often uses standard augmentation as the baseline~\cite{OnMixupTrainThul}, we additionally show the results of all methods when trained using standard augmentation.

Fig.~\ref{fig3_ace} visualizes the calibration performance on the in-distribution test set. 
OMADA shows significant improvements over all datasets and model combinations compared to the baseline and all other methods.

We observe that the ACE of the baseline network is relatively low for some networks (e.g. CIFAR-10 + ResNeXt); this is likely due to the fact that early stopping was used during training.  
Further investigations on this are shown in the appendix Section~\ref{early_stopping_exp}.  
We observe larger performance gains for OMADA for harder datasets such as CIFAR-100 (Fig. ~\ref{fig3c_cifar100}), as well as SVHN (Fig. ~\ref{fig3d_svhn}), where the dataset contains multiple class instances (digits) in the same image, introducing high uncertainty.  

The stability of OMADA's performance across models is remarkable. 
The selected networks range from low capacity networks such as DenseNet, larger networks such as WRN and ResNeXt, as well as a network architecture with multiple dense layers (VGG). 
None of the compared methods achieves such low calibration errors across this diverse set of network architectures and multiple datasets, which demonstrates the benefits of OMADA for model-agnostic calibrated network training.

Fig.~\ref{fig3b_cifar10_acc} shows the accuracy of CIFAR-10 across all models. 
It can be observed that OMADA always improves over the baseline classifier and outperforms all other methods. 
This indicates that the increased calibration performance obtained by OMADA does not come at the expense of a drop in accuracy, but rather significantly increases accuracy. 
This observation is consistent across all studied datasets (Section~\ref{appendix_results}).

\begin{figure*} [!ht]
	\centering
	\begin{subfigure}{.48\textwidth}
		\centering
		\includegraphics[width=0.95\textwidth, clip]{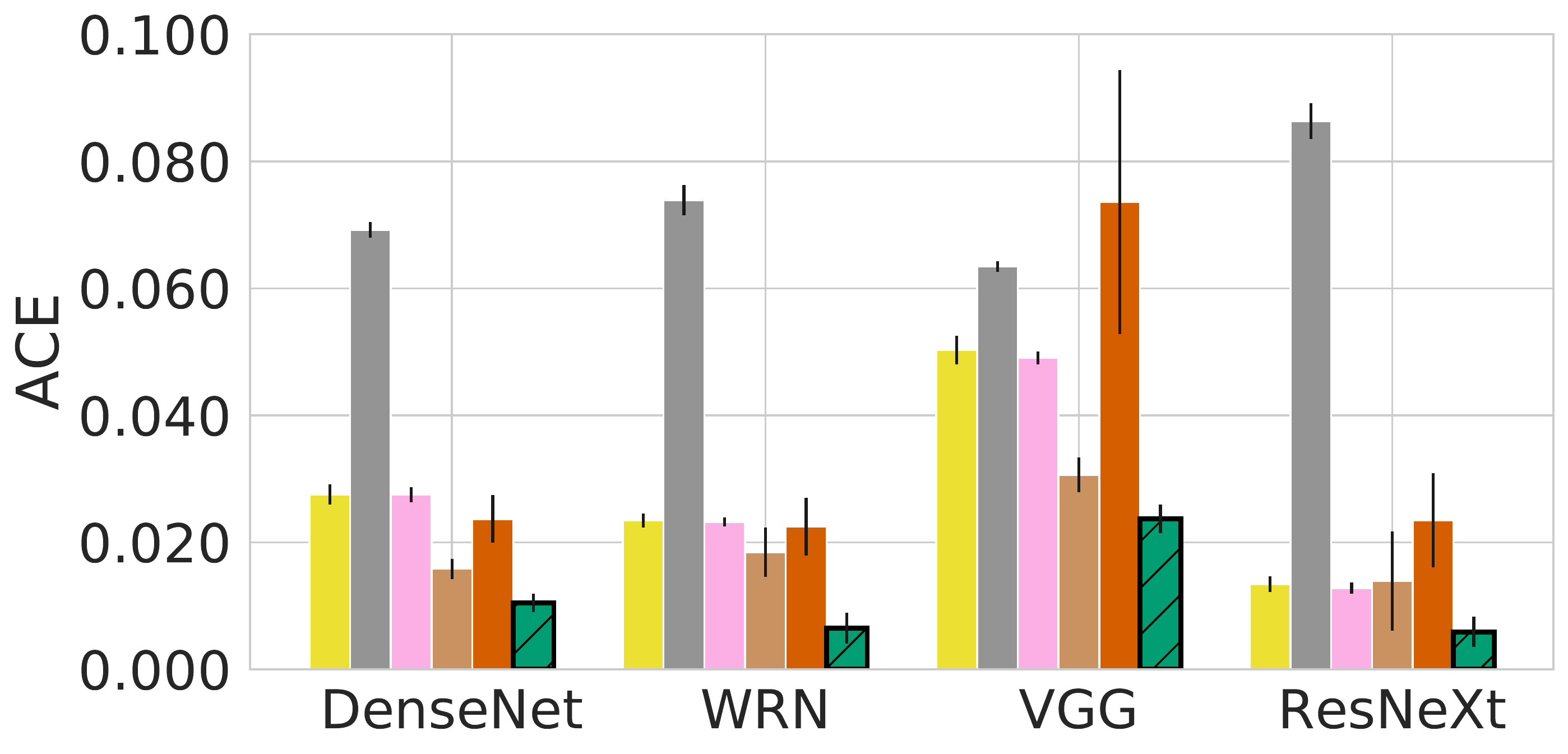}  
		\caption{CIFAR-10 ACE}
		\label{figSA_cifar10_ace}
	\end{subfigure}
	\begin{subfigure}{.48\textwidth}
		\centering
		\includegraphics[width=0.95\textwidth, clip]{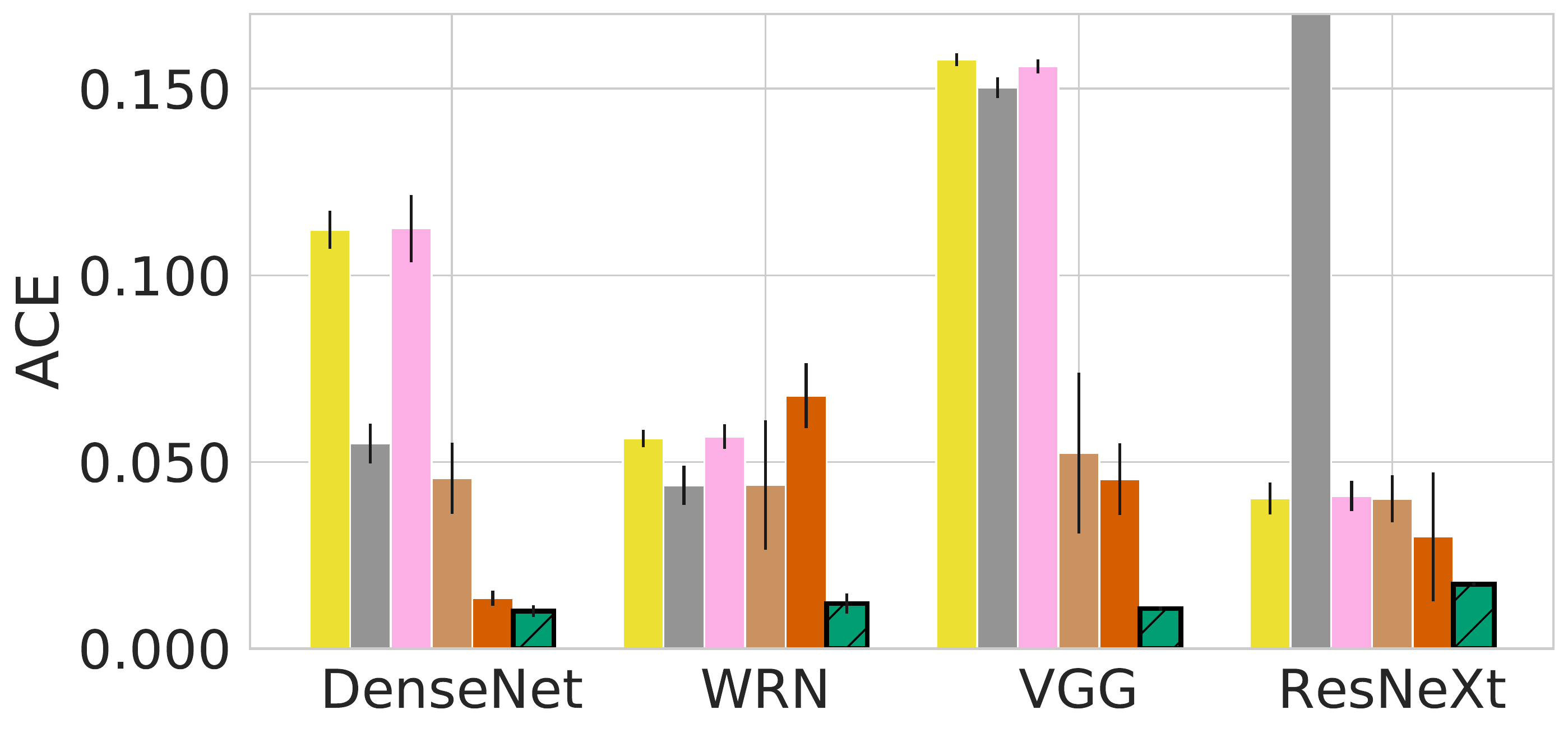}  
		\caption{CIFAR-100 ACE}
		\label{figSA_cifar100_ace}
	\end{subfigure}
	\\
	\begin{subfigure}{.99\textwidth}
		\centering
		\includegraphics[width=0.95\textwidth, clip]{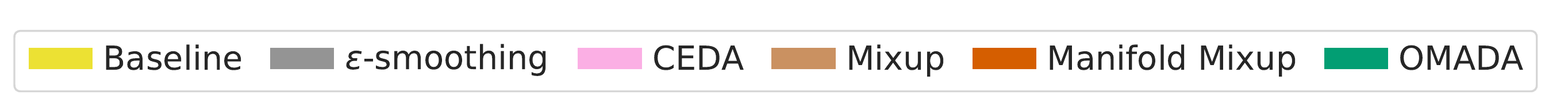}
		\label{legend}
	\end{subfigure}
    \caption{Calibration performance for CIFAR-10 (a) and CIFAR-100 (b) with all methods using standard augmentation during training. Similar to previous observations, OMADA results in a lower ACE than all other calibration methods. This shows that OMADA can be combined with orthogonal generalization techniques to improve accuracy and at the same time yields better calibration. Accuracy plots can be found in Section~\ref{app_xplusSA_acc}}
	\label{figSA_ace_acc}
\end{figure*}

\textbf{Standard Augmentation Baseline}
We visualize the calibration performance using standard augmentation baseline in Fig.~\ref{figSA_ace_acc} (accuracy plots in Fig.~\ref{figAPPENDIX_SA_ace_acc}).
The results are consistent with Fig.~\ref{fig3_ace} and show that OMADA can easily be combined with alternative generalization techniques (thereby boosting accuracy), while maintaining the best calibration performance compared to all other methods.
Interestingly, comparing the observations from the two different baselines, it can be seen that even though standard augmentation improved the accuracy of all methods, the calibration performance has not always stayed the same.
Some methods have significantly higher ACEs, for example, CEDA and Mixup on CIFAR-100 for DenseNet and Manifold Mixup on CIFAR-100 for WRN.
This shows that standard augmentation on its own strongly influences the calibration quality of the networks, thus making it harder to isolate the effect of each calibration method.
Nonetheless, OMADA did not compromise its original calibration gains when using standard augmentation for improving accuracy.
Standard augmentations like crops and flips are specific to images, therefore the comparison of calibration gains in both the base and the augmented setup is important to highlight the strengths of each technique for potential future applications in non-image domains (e.g. text or point clouds).

In summary, for in-distribution samples, OMADA results in well-calibrated, accurate classifiers across all diverse network architectures and datasets, especially in comparison to competing label smoothing/data augmentation approaches.

\begin{figure} [t!]

\centering
\begin{subfigure}{.48\columnwidth}
  \centering
  \includegraphics[width=0.95\textwidth, clip]{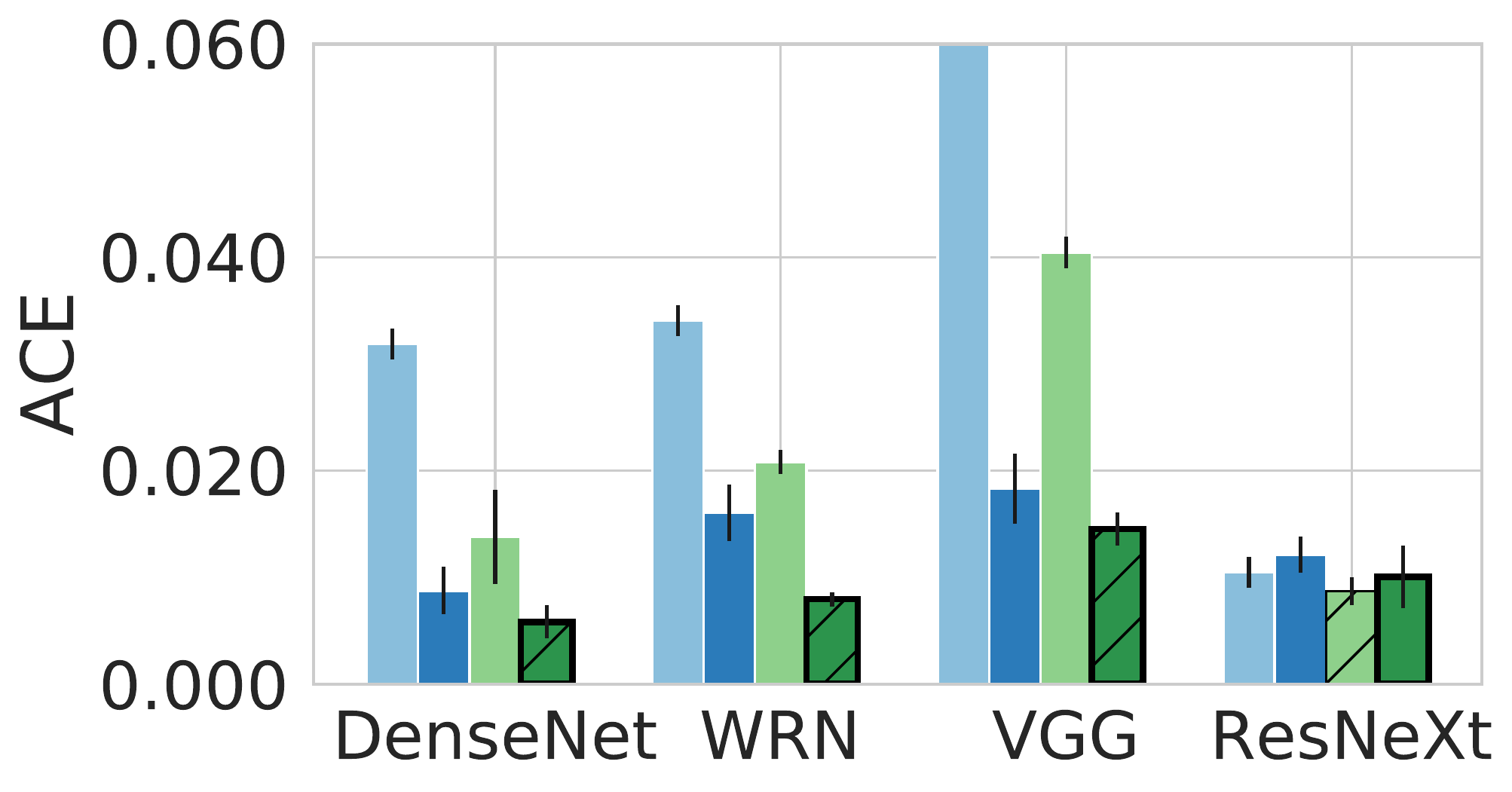}  
  \caption{CIFAR-10}
  \label{fig4a_cifar10_ace}
\end{subfigure}
\begin{subfigure}{.48\columnwidth}
  \centering
  \includegraphics[width=0.95\textwidth, clip]{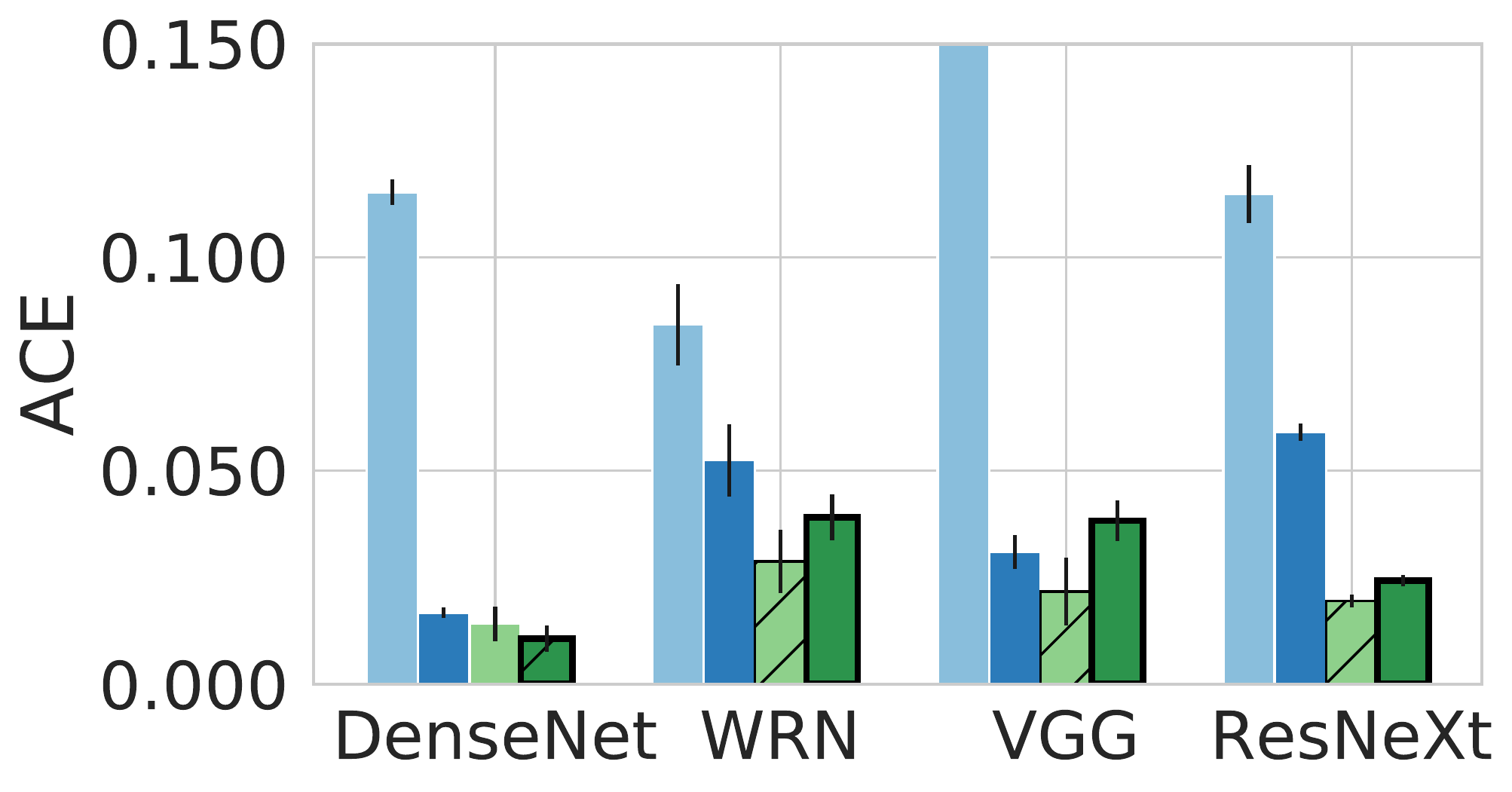}
  \caption{CIFAR-100}
  \label{fig4b_cifar100_ace}
\end{subfigure}
\begin{subfigure}{.99\columnwidth}
  \centering
  \includegraphics[width=0.6\textwidth, clip]{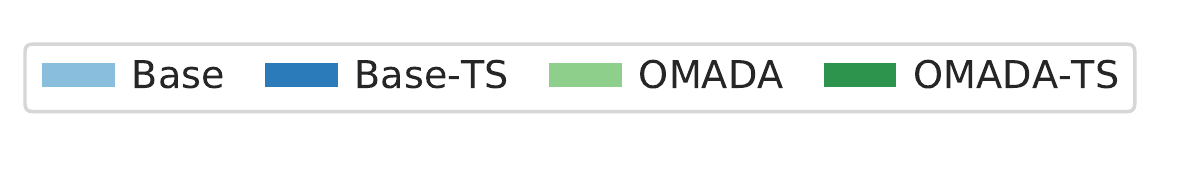}
  \label{legend}
\end{subfigure}
\vspace{-0.5cm}
\caption{Calibration performance (ACE) of OMADA and Temperature Scaling (TS) on in-distribution data. Hatched bars indicate the best-performing method. Error bars are $\pm$ 1 std. dev. over 5 runs. }
\label{fig4_ts_ace}
\end{figure}

\textbf{Temperature scaling} As TS~\cite{GuoPSW17} is an orthogonal post-processing calibration technique, we separately compare the effect of TS applied to the baseline network as well as on OMADA.
Fig.~\ref{fig4_ts_ace} compares the ACE of the baseline and OMADA with their respective TS variants. It can be seen that for CIFAR-10, TS on the baseline mostly surpasses the calibration performance of OMADA alone, but the best performance is obtained by applying TS on top of OMADA. For harder datasets such as CIFAR-100, OMADA alone achieves a similar or often better ACE than Base-TS.

An interesting observation about TS can be seen in Fig.~\ref{fig4_ts_ace}: OMADA-TS does not always produce better calibration when compared to OMADA. This is an unintuitive effect, though further investigation showed similar behavior for other methods in the literature, usually in the case where the calibration error without TS is already fairly low (like in the case of OMADA). This can happen as the NLL for which TS is optimized for, is not directly correlated with the ACE metric. This result calls for careful consideration when using TS for calibration, as it degrades performance for already calibrated networks. A simple alternative is to simply do a grid search over temperatures and choose the one which results in the best calibration performance on a validation set. For further explanation, and an example of this phenomenon, see Section~\ref{temp_scaling_appendix} in the appendix. Furthermore, as TS does not change the accuracy of the models, it does not come with the accuracy improvements of OMADA.  

\textbf{OMADA Variants}
Here we show the performance of different OMADA variants, to investigate the effects of adding ambiguous images and soft labels independently.  
We first investigate an alternative sampling method, which preferentially samples images from the path with high label entropies (i.e. higher chance of sampling pure boundary region samples), as opposed to uniformly sampling them. We call this variant OMADA-SE (\emph{Sample from Entropy}). Furthermore, we study the effect of the soft labels produced by the latent-space classifier by training the networks with the generated ambiguous samples from OMADA and OMADA-SE, but changing the labels.
We either harden the soft labels based on the maximum class probability (OMADA*-H), or change the class labels to be uniform across all classes (OMADA*-U).
We investigate the resulting network calibration (ACE), the accuracy (ACC), and the outlier detection performance (AUC) of the variants. The results are shown in Table~\ref{table1_ablations} for CIFAR-10 on DenseNet and WRN.

\begin{table}[t!]
    \begin{center}
	\captionsetup{singlelinecheck=false}
	\begin{tabular}{lrrrrrr}
		\toprule
		CIFAR-10 & \multicolumn{3}{c|}{DenseNet}   &  \multicolumn{3}{c}{WRN} \\
		&     ACE &     ACC &      \multicolumn{1}{c|}{AUC} &    ACE &     ACC &     AUC \\
		\midrule
		 Base &  0.0319 &  92.768 &  \multicolumn{1}{r|}{0.9076}  &  0.0341 &  91.596 &  0.9022 \\
		 OMADA &  0.0138 &  94.530 &   \multicolumn{1}{r|}{0.9252}&  0.0208 &  95.772 &  0.9243 \\
		OMADA-H &  \textbf{0.0058} & 93.652  &    \multicolumn{1}{r|}{0.8913} &  \textbf{0.0172} &  95.150 &  0.9210 \\
		OMADA-U &  0.0260 &  93.890 &  \multicolumn{1}{r|}{0.9678} &  0.0274 &  95.446 &  0.9750 \\
		 OMADA-SE &  0.0273 &  \textbf{94.988} &  \multicolumn{1}{r|}{0.9741} &   0.0207 &  \textbf{96.022} &  \textbf{0.9833} \\
		 OMADA-SE-H &  0.0281 &  94.258 &  \multicolumn{1}{r|}{0.9119} &  0.0222 &  95.882 &  0.9643 \\
		 OMADA-SE -U &  0.0302 &  94.618 &   \multicolumn{1}{r|}{\textbf{0.9786}} &  0.0231 &  95.248 &  0.9346 \\
		\bottomrule
	\end{tabular}
	 \end{center}
	\caption{Performance of OMADA ablation methods on calibration (ACE), network accuracy (ACC), and outlier detection (AUC). -H refers to the respective hard label variant, -U refers to the respective uniform label variant. DN refers to DenseNet.  Bold entries indicate the best-performing method. We report the mean over $5$ independent runs for each method (std devs. can be found in the appendix).}
	\label{table1_ablations}
   
\end{table}

We observe in Table~\ref{table1_ablations} that using this alternative sampling method performs very competitively on multiple tasks, especially on outlier detection. 
The effect of hardening the labels yields surprisingly good results on ACE, where it sometimes improves calibration over the corresponding soft label variant, suggesting that the ambiguous images generated by the on-manifold attacks alone are enough to improve the network's confidence estimates. 
However, this gain comes at the cost of a drop in accuracy, suggesting that the soft labels help generalization. This observation will be the focus of future research.

The effect of hardening labels is different for the sampling variants; as OMADA-SE contains more samples with higher entropy soft labels, the change in label density is much more drastic than in OMADA, which also produces samples far away from decision boundaries (i.e. the soft label is already relatively hard).  This is illustrated especially in the outlier detection performance: here, OMADA-H and OMADA-SE-H suffer in comparison to their soft-label counterparts. These observations are consistent with OOD-MMC reported in the appendix (Section~\ref{ablations_appendix}).

\begin{figure*} [htp]
	\centering
	\begin{subfigure}{.48\textwidth}
		\centering
		\includegraphics[width=0.95\textwidth, clip]{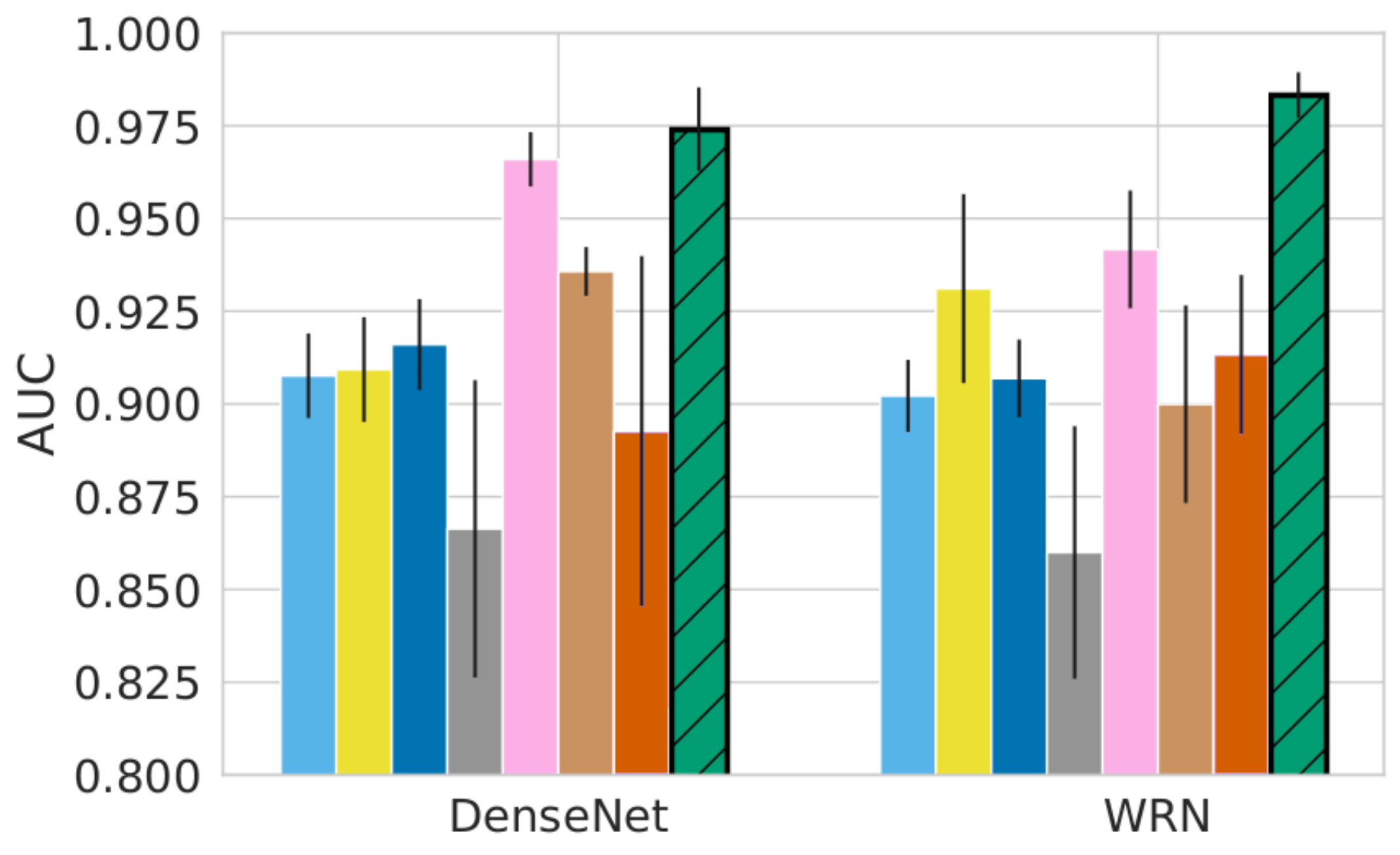}  
	\end{subfigure}
	\begin{subfigure}{.48\textwidth}
		\centering
		\includegraphics[width=0.95\textwidth, clip]{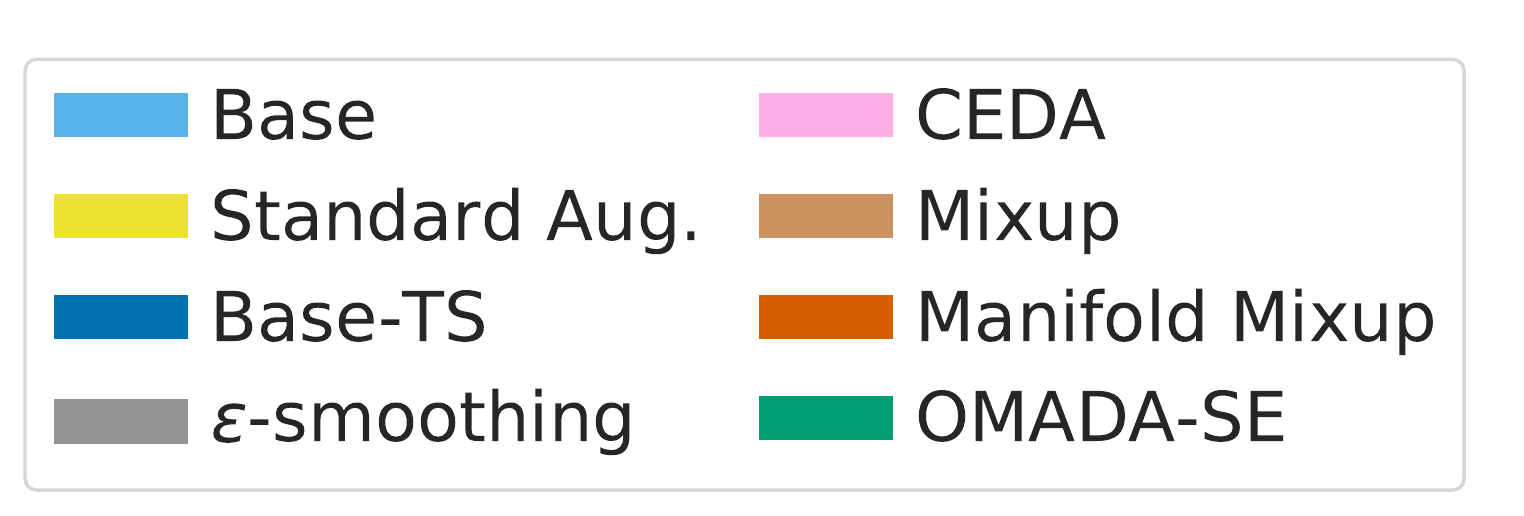}
	\end{subfigure}
    \caption{Outlier detection performance(AUC) of label smoothing methods. Hatched bars indicate the best-performing method. Error bars are $\pm$ 1 std. dev. over 5 runs. }
    \label{fig5_auc}
\end{figure*}

Next, we study the effect of assigning  uniform class labels for each ambiguous sample generated by the adversarial attack. 
The results show that the soft labels of the ambiguous samples are required to attain competitive ACE and accuracy for in-distribution data. However, for out-of-distribution samples, where the AUC and OOD-MMC metrics are optimized when predicting near uniform class labels on OOD data, the OMADA*-U networks do very well. This is consistent with observations from CEDA \cite{ReLUMHein}, where uniform class labels are also used to improve detecting OOD samples (shown in Fig.~\ref{fig5_auc}).

In summary, changing the soft labels increases performance on some tasks, but degrades performance across other tasks; the best choice of labels is then task-dependent. On average, the soft labeled methods (OMADA and OMADA-SE) perform stably across tasks.

\textbf{Outlier Detection} In order to put the outlier detection abilities of OMADA-SE (the best variant across multiple tasks) into context, we compare the AUC to the already-investigated label smoothing methods (Fig.~\ref{fig5_auc}). OMADA-SE outperforms all other methods on both DenseNet and WRN, albeit with a small gap to CEDA on DenseNet. The good performance of CEDA is not surprising, as it is implicitly trained to predict lower confidence on out-of-distribution samples (in CEDA these are random noise images). 
Interestingly, soft-labels alone are not enough to result in good outlier detection, as evidenced by the poor performance of $\epsilon$-smoothing.

\textbf{Stochastic DNN methods}
We compare the calibration and outlier detection performance of the OMADA variants to both Monte Carlo (MC) Dropout \cite{DropoutGalG16} and Ensembles, as these are commonly used to obtain uncertainty estimates, and have been shown to improve network calibration (results in Section ~\ref{stochastic_approximations}).  
As with TS, these methods are both orthogonal to OMADA, and can be easily combined.

\section{Conclusion}
In this article we have introduced the concept of on-manifold adversarial data augmentation for uncertainty estimation by leveraging recent advances in generative modeling. 
By combining a latent space classifier on the approximated data manifold with on-manifold adversarial attacks we derive a novel sampling procedure, which generates samples specifically in challenging regions close to decision boundaries.
The OMADA dataset is generated by using the decoder network to project back into the image space, and using soft labels derived from the latent space classifier.
Through a range of carefully chosen experiments, we study the effect of OMADA as a data augmentation method for training an independent image space classifier.

An extensive set of experiments show significant improvements across multiple datasets and diverse network architectures, as well as on multiple tasks.
The stability of the OMADA results for ACE across multiple networks is a particularly desirable property, as most alternative methods fail to perform well across all investigated networks.
OMADA can be combined with post-processing methods such as temperature scaling \cite{GuoPSW17}, and we
are confident that further beneficial combinations and extensions of the key concept of OMADA will be discovered in future research.
Furthermore, we show that OMADA always results in increased classification accuracy compared to baselines with and without data augmentation, and all other competing methods. 
Finally, OMADA-SE is presented as a method to focus on data generation in boundary regions, thereby outperforming all other methods for outlier detection.

This is a first step towards improving uncertainty quantifications for deep networks through on-manifold adversarial samples. 
Initial results show significant improvements of the networks ability to assign confidence to its predictions on in-distribution samples as well as out-of-distribution samples. Further studies are required to investigate the behavior of these networks on data which marginally leaves the data manifold (e.g. unseen transformations or corruptions).

\newpage
{\small
	\bibliographystyle{ieee_fullname}
	\bibliography{literature}
}
\newpage
\cleardoublepage
\section*{Appendix: On-manifold Adversarial Data Augmentation Improves Uncertainty Calibration} \large
\renewcommand{\thetable}{A\arabic{table}}
\renewcommand{\thefigure}{A\arabic{figure}}
\renewcommand{\thesection}{A\arabic{section}}
\setcounter{table}{0}
\setcounter{figure}{0}
\setcounter{section}{0}
\setcounter{page}{1}

\section{Visualizing Other Attack Paths}
\begin{figure*}[!ht]
  \includegraphics[width=1.0\linewidth]{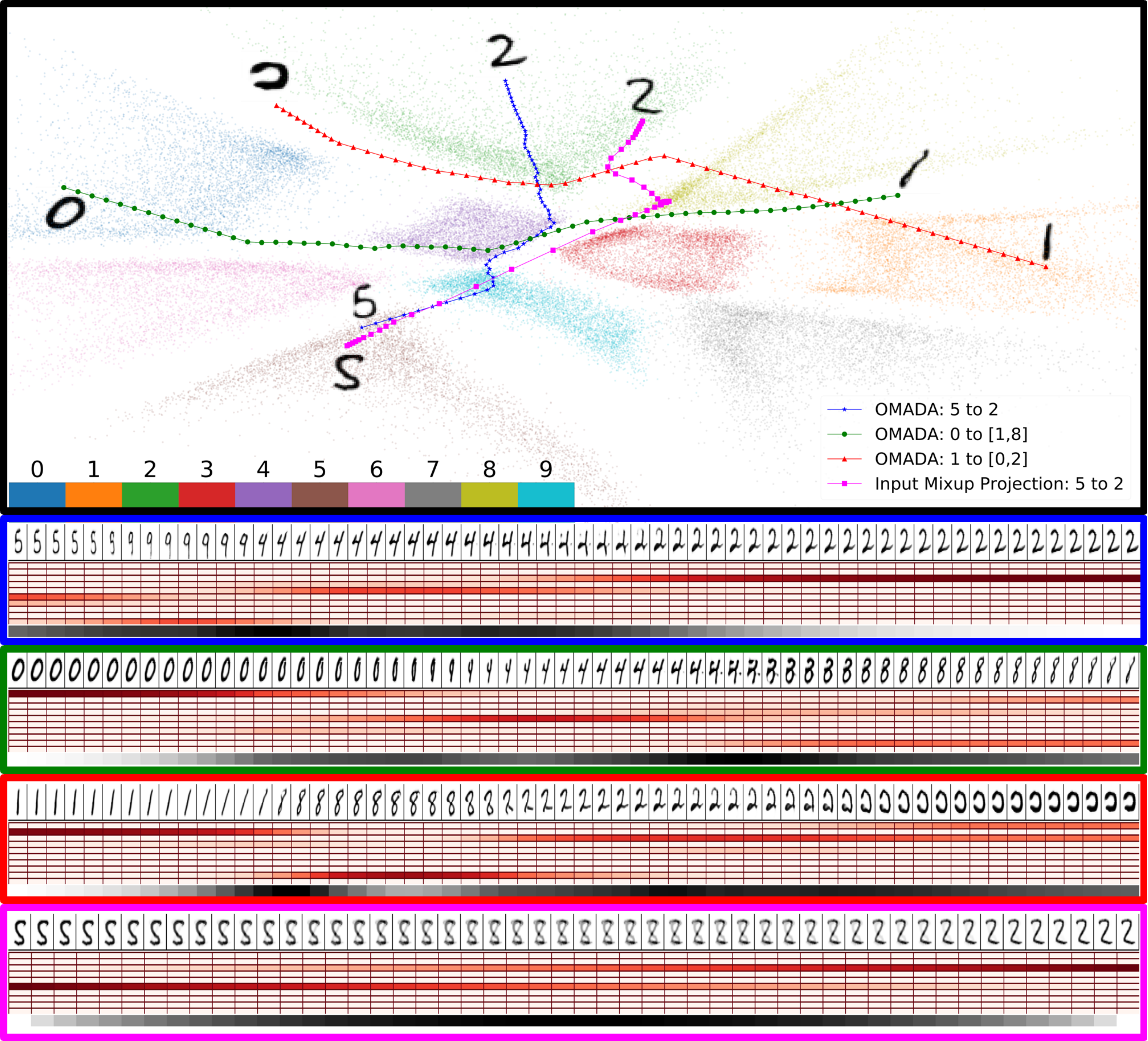}
  \caption{Visualization of an MNIST encoder-decoder latent space with multiple trajectories traversing through the latent space. The paths depict $3$ On-manifold adversarial attack paths, as well as $1$ Input Mixup projection into the same latent space. Below the latent space we visualize the decoded image path for OMADA (top 3 blocks) and the Input-Mixup images (bottom block) along with their corresponding soft labels (10 rows below images, red intensity corresponds to likelihood for classes 0 to 9), and the class entropy (bottom row, black shows high entropy). 
  The green and red paths are generated when setting the target as a soft-label between two classes (targeting specifically the decision boundaries). For example, the green path starts at cluster ``0" and optimizes Eq.~\ref{eq:hardtgt} with the target ($y_i^{o}$) set to a soft label with $0.5$ for classes $1$ and $8$, and $0$ elsewhere. As a result, this produces perturbations which direct the path to the decision boundary between the classes $1$ and $8$.  
  The magenta path shows the projection of Input Mixup images between samples ``5" and ``2".
  It can be seen the OMADA paths produce mostly confusing samples at the decision boundaries, and that the soft labels reflect this confusion, whereas Input Mixup produces images which resemble an ``8" (seen in the image path below as well as the magenta projection path going to the ``8" cluster first before heading towards the target cluster `2"); Mixup's soft label does not reflect this in its soft-label (soft-label is zero at class ``8").   }
  \label{Afig_a1_latent_paths}
\end{figure*}

Fig.~\ref{Afig_a1_latent_paths} depicts more examples of attack paths, with different start and end targets, produced by the presented method.
The OMADA attack path examples include paths where the target is set to another class (e.g. blue path with target ``2"), as well as paths where the target is a decision boundary (e.g. green path with target between ``1" and ``8" and red path with target between ``0" and ``2").
The decision boundary between two classes can be reached by setting the target vector ($y_i^{o}$) in Eq.~\ref{eq:hardtgt} to $0.5$ for the two classes and $0$ elsewhere.
It can be seen that the images produced by the decision boundary paths produce confusing samples which reflect features from the neighboring clusters.
Furthermore, this confusion is reflected in the soft-label.

\section{Input Mixup Example} \label{appendix_interpolation_methods}
In Fig.~\ref{Afig_a1_latent_paths}, the Input Mixup projection path is visualized in magenta.
This path is produced by projecting the linearly interpolated images produced by Input Mixup into the latent space using the encoder.
Even though Mixup mainly produces unrealistic images (Fig.~\ref{fig:teaser}), when it does produce realistic samples from another class, the soft label would not reflect the presence of this class.
For example, in Fig.~\ref{Afig_a1_latent_paths}, Input Mixup produces an interpolated image between the classes ``5" and ``2" which looks similar to an ``8". It can be seen that Input Mixup assigns zero probability for class ``8",  whereas using our encoder the images get mapped to the ``8" cluster, which means a soft label produced by our method would reflect the presence of the class ``8" .

\section{Experiment Hyperparameters} \label{experiment_hyperparameters}
This section will present detailed information regarding the training process.

\subsection{Model and Training Hyper-parameters}
All optimizer training hyper-parameters for the training of the image-space classifiers can be found in Table~\ref{tab:training_info}. 
These parameters are kept unchanged across the three datasets and all methods, as well as across all $5$ repetitions (where only the random seed was changed). 

\begin{table}[!ht]
\small
    \begin{center}
\begin{tabular}{ |l|c|c|c|c|c|c|c|c|c|c|c| } 
\hline
{Model} & {Num Params.} & {Weight} & {Epochs} & {LR} & {Milestones} & {LR} & {Batch} \\
{} & {(10 classes)}  & {Decay} & {} & {Scheduler} & {} & {Decay} & {Size} \\
\hline
\hline
    DenseNet-100-12~\cite{dnpaper} & 796, 162 &            \num{1e-4} & $300$ & Multi-step & $[150,255]$ & $0.1$ & 64 \\     
    WRN-28-10~\cite{wrnpaper} & 36, 479, 194 &                                          \num{5e-4} & $200$ & Multi-step & $[60,120,160]$ & $0.1$ & 128 \\   
    VGG-16~\cite{vggpaper} & 33, 646, 666 &                     \num{1e-4} & $160$ & Multi-step & $[80,120]$ & $0.1$ & 128 \\
    ResNeXt-29~\cite{resnextpaper} & 34, 426, 698    &                                     \num{5e-4} & $300$ & Multi-step & $[150,255]$ & $0.1$ & 128 \\  
\hline
\end{tabular}
    \end{center}
\caption{Training hyper-parameters. We use SGD with a base learning rate of $0.1$ and momentum of $0.9$ for all trainings. }
\label{tab:training_info}
\end{table}

\subsection{OMADA training hyper-parameters}
Each OMADA-trained network uses a balanced $50\%$ of real training samples (with hard one-hot labels) and $50\%$ of the On-manifold adversarial samples in each \emph{batch}.
In order to enable direct comparison to alternative methods in the literature, we ensure that for each epoch, the total number of gradient updates performed are the same with the balanced number of samples from both datasets.
Therefore, at the end of each epoch, $50\%$ of the real training samples are not seen and instead replaced by On-manifold adversarial samples. 
It should be noted that the $50\%$ real samples seen during each epoch vary across epochs. 
In order to speed up the training process, we create an offline On-manifold adversarial dataset, and sample from this dataset to fill up each batch during training.
\\
For all networks, $1$K random training samples are withheld to create a validation set. 
The validation set accuracy is used for early stopping, and all experiments (unless stated otherwise) report the results from the checkpoint with the highest validation accuracy during training.
Furthermore, the validation set is used to find the best temperature to produce the temperature scaling results.
\\

\subsection{Details about Literature Methods}
This sub-section reports the hyper-parameters of the alternative methods in the literature in more detail.
\begin{enumerate}
	\item \textbf{Base}: base network trained using only real samples with hard labels and no data augmentation.
	\item \textbf{Standard augmentation (std\_aug)}: base network trained with data augmentation on the training samples (random crop (padding$=4$) and horizontal flips (flip prob.$=0.5$)).
	\item \textbf{Mixup}: mixup training~\cite{Zhang2018mixupBE} with $\alpha=0.1$~\cite{OnMixupTrainThul}. Augments the training dataset by linearly interpolating between both images and labels within a mini batch.
	\item \textbf{Manifold Mixup}: extends mixup training by taking linear interpolations of hidden layers in the network and hard labels. We use $\alpha=2.0$~\cite{manifoldmixup}.
	\item \textbf{$\epsilon$-smoothing}: smooths the labels with $\epsilon=0.1$ (found to be best in~\cite{OnMixupTrainThul}) by taking a linear combination of the $(1-\epsilon) \times \text{hard-label}$ and $\epsilon \times \text{uniform-class-label}$.
	\item \textbf{CEDA}: confidence enhancing data augmentation (CEDA) is a training scheme that enforces uniform confidences on out-of-distribution noise. These out-of-distribution images are included into the training by replacing half of the batch of real samples with $25\%$ permuted pixel images and $25\%$ uniform random noise images. For each of these augmented images, a Gaussian filter with standard deviation $\sigma \in [1.0, 2.5]$~\cite{ReLUMHein} is applied on the images, to have more low-frequency structure in the noise. The label for each of these images is the uniform class label.
\end{enumerate}

\section{Additional Results} \label{appendix_results}
\subsection{Classification Accuracy}
\begin{figure*}[htp]
	\centering
	\begin{subfigure}{.48\textwidth}
		\centering
		\includegraphics[width=0.95\textwidth, clip]{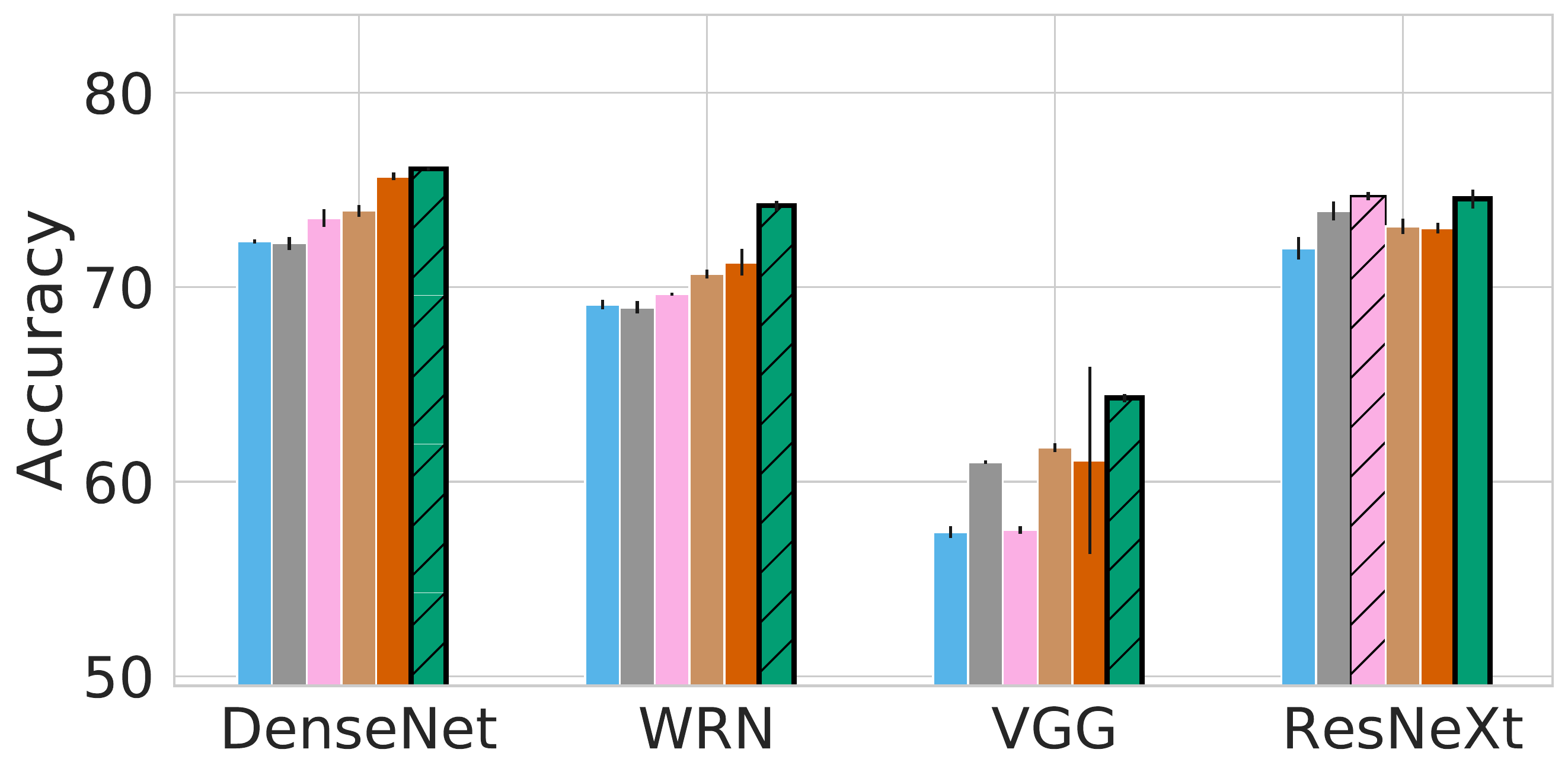}  
		\caption{CIFAR-100 Accuracy}
		\label{Afig1_acc_cifar100}
	\end{subfigure}
    \hfill
	\begin{subfigure}{.48\textwidth}
		\centering
		\includegraphics[width=0.95\textwidth, clip]{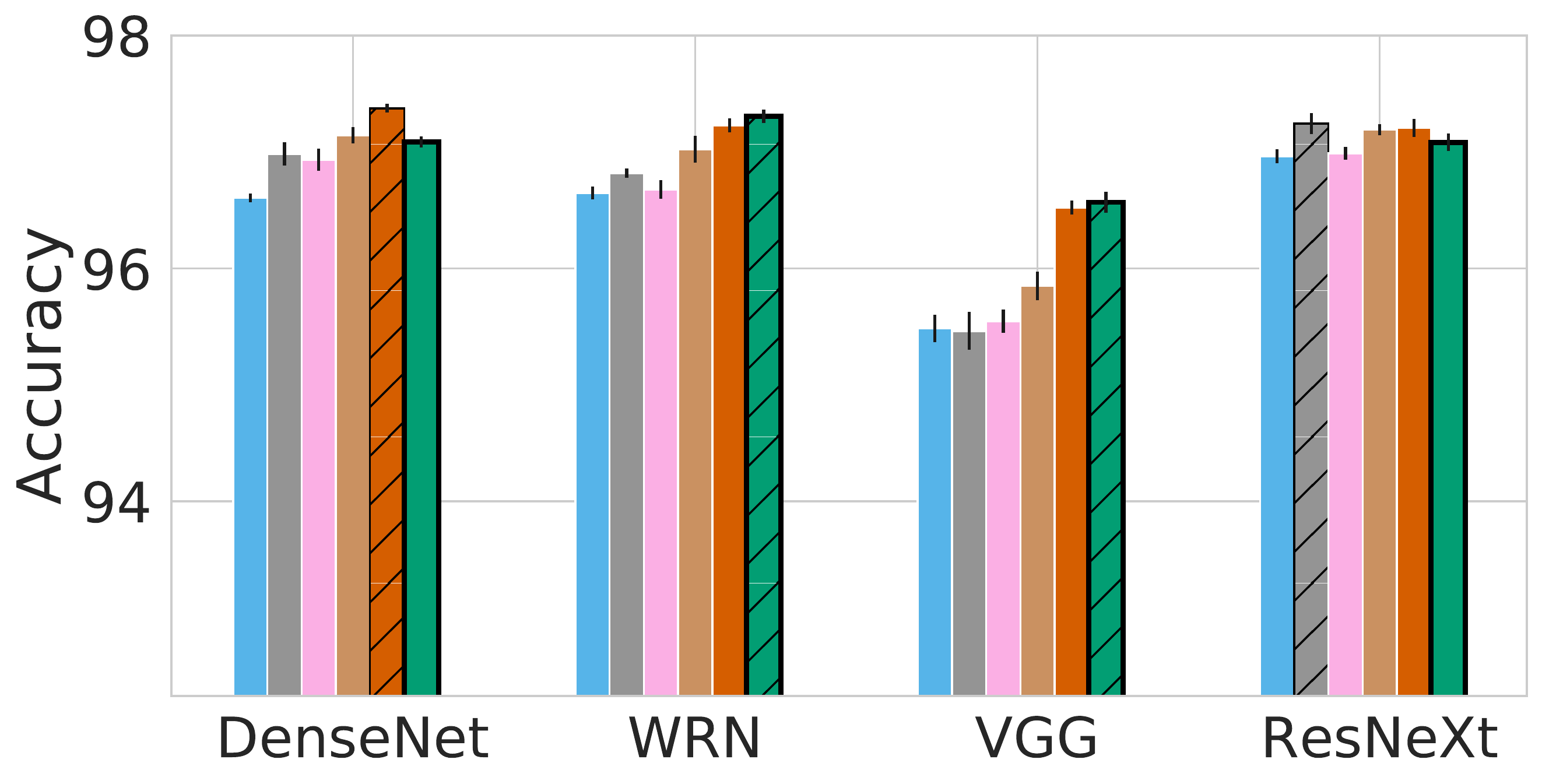}
		\caption{SVHN Accuracy}
		\label{Afig1_acc_svhn}
	\end{subfigure}
		\begin{subfigure}{.99\textwidth}
		\centering
		\includegraphics[width=0.95\textwidth, clip]{newdata/main_paper_figures/orig_figure_legend.pdf}
		\label{legend}
	\end{subfigure}
	\caption{Classification accuracy on CIFAR-100 (a) and SVHN (b). Across both datasets, we observe similar behavior as~\ref{fig3b_cifar10_acc}. Error bars are $\pm$ 1 std. dev. over 5 runs. }
	\label{AfigACCURACY}
\end{figure*}

In Fig.~\ref{AfigACCURACY} we report the classification accuracy for CIFAR-100 and SVHN. 
We make similar observations as in Fig.~\ref{fig3b_cifar10_acc}.
OMADA achieves an improvement in accuracy compared to the Base models and most other methods, emphasizing that the gain in calibration does not come at the cost of a drop in accuracy compared to the baseline network without OMADA. 
\\

\subsection{Classification Accuracy of Networks Trained with Standard Augmentations}\label{app_xplusSA_acc}
\begin{figure*} [!ht]
	\centering
	\begin{subfigure}{.48\textwidth}
		\centering
		\includegraphics[width=0.95\textwidth, clip]{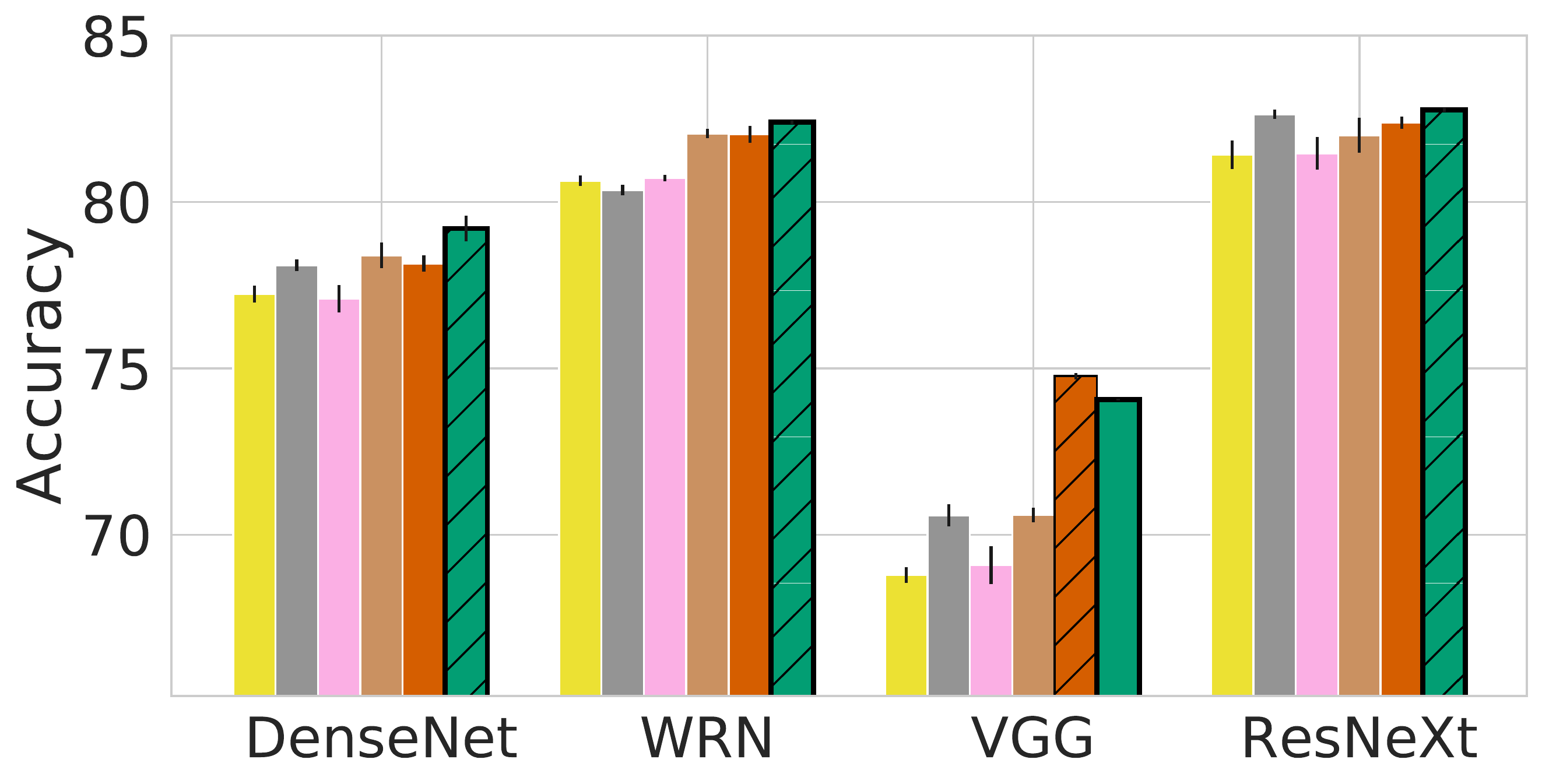}
		\caption{CIFAR-10 Accuracy}
		\label{figSA_cifar10_acc}
	\end{subfigure}
	\hfill
	\begin{subfigure}{.48\textwidth}
		\centering
		\includegraphics[width=0.95\textwidth, clip]{newdata/main_paper_figures/accuracy_new_cifar100.pdf}
		\caption{CIFAR-100 Accuracy}
		\label{figSA_cifar100_acc} 
	\end{subfigure}
	\\
	\begin{subfigure}{.99\textwidth}
		\centering
		\includegraphics[width=0.95\textwidth, clip]{newdata/main_paper_figures/new_figure_legend.pdf}
		\label{legend}
	\end{subfigure}
    \caption{ Classification accuracy on CIFAR-100 (a) and SVHN (b). Accuracy for CIFAR-10 (a) and CIFAR-100 (b) with all methods using the standard augmentations baseline during training. Results are similarly consistent with previous observations which do not use standard augmentations. Error bars are $\pm$ 1 std. dev. over 5 runs.}
	\label{figAPPENDIX_SA_ace_acc}
	\vspace{-0.5cm}
\end{figure*}

In Fig.~\ref{figAPPENDIX_SA_ace_acc} we report the classification accuracy for CIFAR-10 and CIFAR-100 for all methods using standard augmentations (random crops and horizontal flips).
Similar to the baseline networks without standard augmentations, we observe that OMADA consistently produces more accurate networks than the baseline and most other methods.
In combination with the calibration results of the networks which use standard augmentations (Fig.~\ref{figSA_ace_acc}), we show that OMADA can be combined with orthogonal generalization methods and still maintain low calibration errors as well as improve accuracy compared to the baseline.

\subsection{Sparsification and OOD-MMC}
In this section we report the Sparsification results on the in-distribution test set and Mean Maximal Confidence on out-of-distribution data (OOD-MMC). 
Sparsification evaluates how well a given uncertainty estimate correlates with the true error; intuitively, we want our networks to be more confident about correct predictions, and less confident about incorrect predictions \cite{geifman2017coverage}.  
This is calculated by selectively calculating the classification accuracy on increasingly large subsets of the test set.  
Samples are added to the subset based on their uncertainty; the more certain samples are added first. 
The final metric is the difference between the curve generated by the method and the ideal curve, in which all incorrectly-classified images have a higher uncertainty than all correctly-classified images. 
A lower Sparsification error is desired.

Another measure for evaluating the over-confidence of networks is to measure the OOD-MMC on out-of-distribution data. 
For out-of-distribution samples we want the network to assign a confidence of $\frac{1}{\text{\# classes}}$, reflecting maximum uncertainty. 
The Mean Max Confidence (MMC) measures how well the network performs the task of assigning a low confidence to unseen samples.

\begin{figure*} [htp]
	\centering
	\begin{subfigure}{.48\textwidth}
		\centering
		\includegraphics[width=0.95\textwidth, clip]{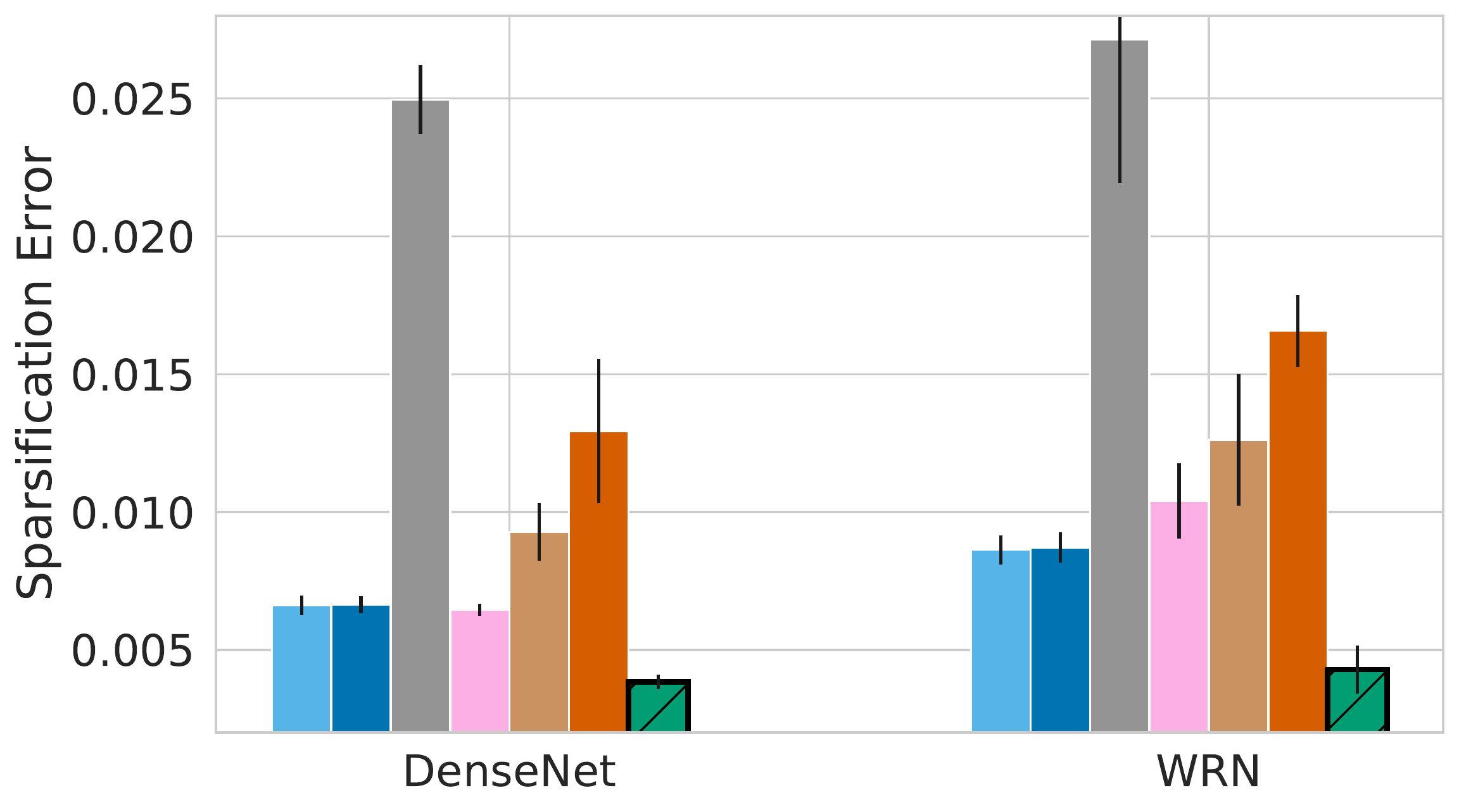}  
		\caption{CIFAR-10 Sparsification Error}
		\label{Afig2_sparsification_cifar10}
	\end{subfigure}
	\begin{subfigure}{.48\textwidth}
		\centering
		\includegraphics[width=0.95\textwidth, clip]{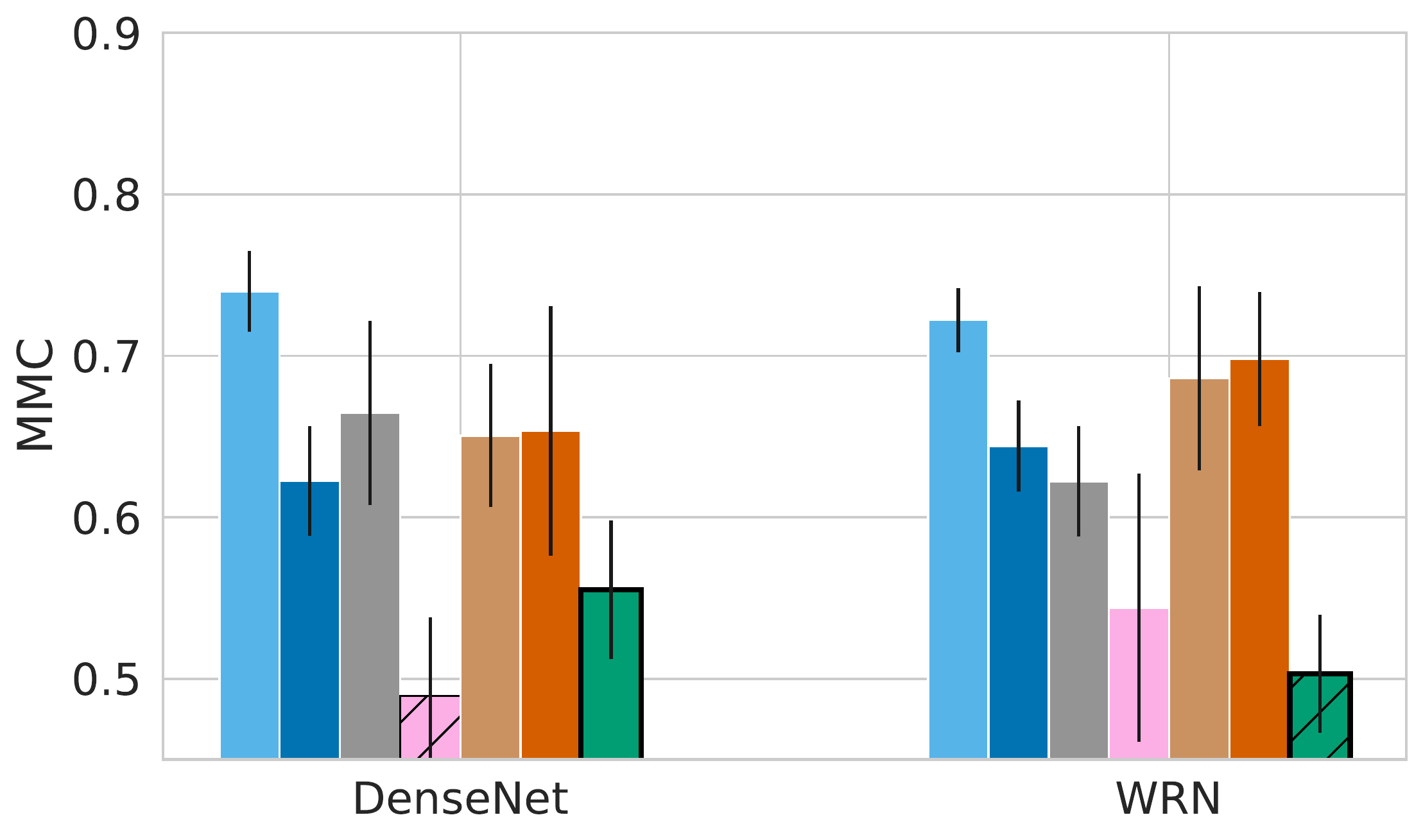}
		\caption{CIFAR-10 OOD-MMC}
		\label{Afig2_mmc_cifar10}
	\end{subfigure}
		\begin{subfigure}{.99\textwidth}
		\centering
		\includegraphics[width=0.95\textwidth, clip]{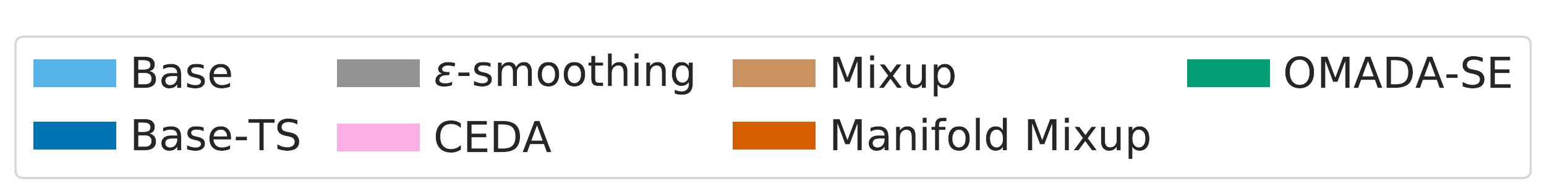}
		\label{legend}
	\end{subfigure}
	\caption{Sparsification (a) and OOD-MMC (b) for CIFAR-10 on DenseNet and WRN. For Sparsification error we observe that OMADA-SE has a significantly lower error compared to all other methods except Standard Augmentation. For OOD-MMC, we observe that the best performing methods are CEDA and OMADA-SE. Both methods have significantly lower confidence assigned to OOD data compared to all other methods. Error bars are $\pm$ 1 std. dev. over 5 runs. }
	\label{AfigSPARSIFICATION_MMC}
\end{figure*}

These results can be seen in Fig.~\ref{AfigSPARSIFICATION_MMC}. 
We observe that OMADA-SE significantly improves its Sparsification error compared to all other methods, and performs similar to Standard Augmentation. 
For OOD-MMC, we observe that OMADA-SE performs better than all other methods except CEDA, which has the lowest OOD-MMC on DenseNet. 
As CEDA augments the training dataset with out-of-distribution images and uniform class labels, it has an advantage for the OOD-MMC metric which is optimized by predicting high entropy labels (i.e. uniform class labels).
Though for WRN, OMADA-SE again performs best compared to all other methods.
\\

\section{ACE on Last Epoch Checkpoint} \label{early_stopping_exp}
\begin{figure}[ht]
\centering
\begin{subfigure}{1\columnwidth}
  \centering
 \includegraphics[width=1.0\linewidth]{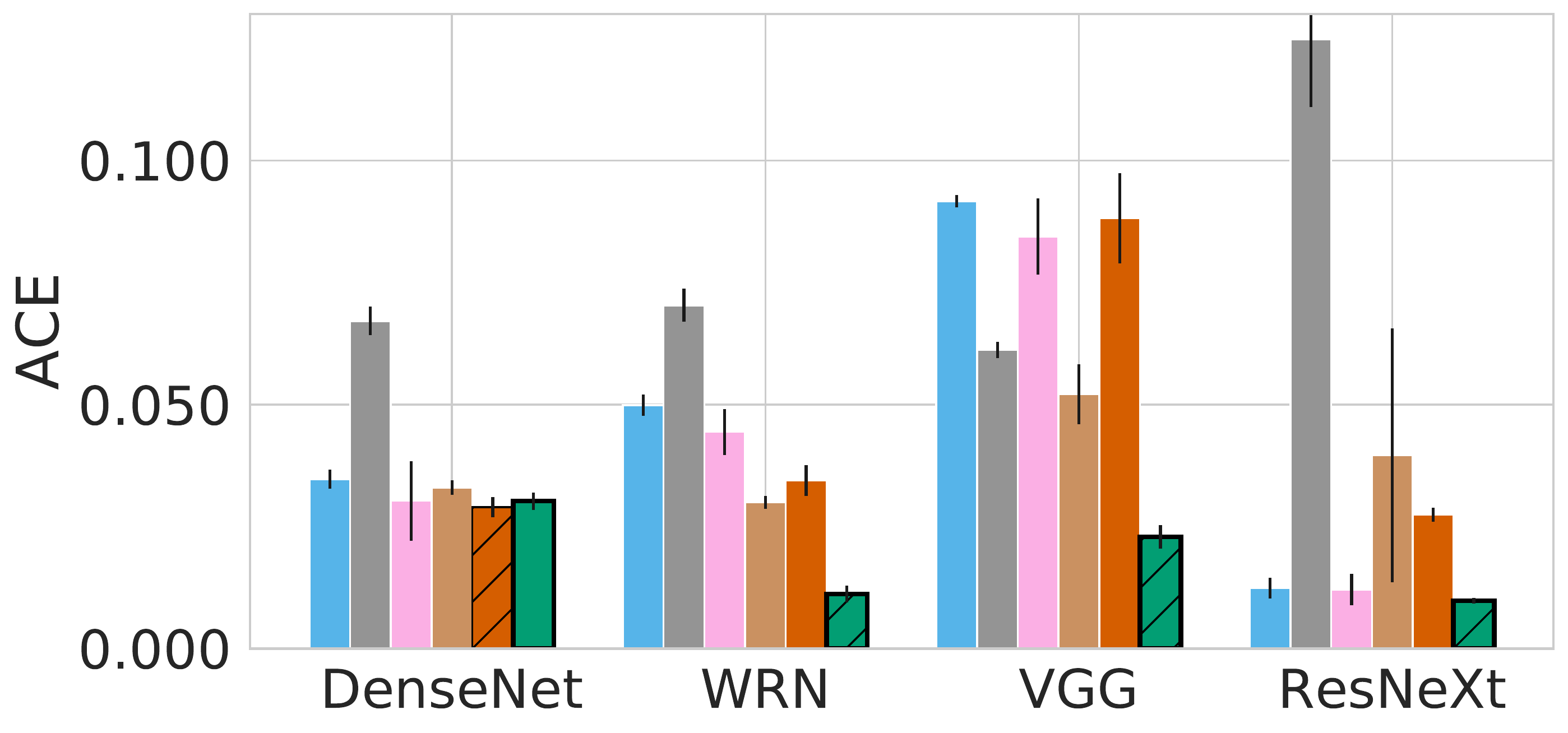}
\end{subfigure}
\\
\begin{subfigure}{1\columnwidth}
  \centering
  \includegraphics[width=0.99\linewidth, clip]{newdata/main_paper_figures/orig_figure_legend.pdf}
  \label{legend}
\end{subfigure}
\caption{Calibration performance (ACE) of label-smoothing methods on in-distribution test data for CIFAR-10 for the last epoch checkpoint. Hatched bars indicate the best-performing method per network architecture Error bars are $\pm$ 1 std. dev. over 5 runs..}
\label{Afig3_ace_cifar10_last_epoch}
\end{figure}

In the main paper, we report the evaluations based on the model weights resulting in the highest validation accuracy. 
In order to show that the results are consistent with results from the last epoch model weights, in this section we report the ACE results for CIFAR-10 on all models. 
This ensures that all models were trained for the exact same number of epochs.
Fig.~\ref{Afig3_ace_cifar10_last_epoch} shows ACE results for CIFAR-10 for all models for the last epoch checkpoint. 
It can be seen that similar performance orderings can be observed compared to Fig.~\ref{fig3a_cifar10_ace}. 
Most methods have a worse ACE when evaluating using the last epoch (as longer training often increases mis-calibration), though surprisingly some exceptions do exist.
This suggests a further study into the temporal aspect of network calibration across training epochs would be informative. 

\section{Ablation Studies} \label{ablations_appendix}
Here we report the OOD-MMC results of the ablation study, as well as the standard deviations of the networks in Table~\ref{table1_ablations}.
We report these numbers in Table~\ref{appendix_table_ablations}.
It can be seen that similar to AUC, the soft-labels become important to get a lower MMC on out-of-distribution data.
\begin{table*}[t!]
   \centering
	\captionsetup{singlelinecheck=false}
\begin{tabular}{p{10mm}lcccc}
\toprule
    \multicolumn{2}{l}{CIFAR-10}  &                ACE &                ACC &                AUC &                OOD-MMC \\
\midrule
\multicolumn{1}{l|}{DN} & Base &  $0.0319\pm0.0015$ &  $92.768\pm0.1425$ &  $0.9076\pm0.0114$ &  $0.7399\pm0.0251$ \\
    \multicolumn{1}{l|}{} & OMADA &  $0.0138\pm0.0044$ &  $94.53\pm0.1349$ &  $0.9252\pm0.0209$ &  $0.6799\pm0.0468$ \\
    \multicolumn{1}{l|}{}	& OMADA-H &  $\mathbf{0.0058\pm0.002}$ &  $93.652\pm0.1766$ &  $0.8913\pm0.0243$ &  $0.7045\pm0.0362$ \\
    \multicolumn{1}{l|}{}	& OMADA-U &  $0.026\pm0.0023$ &  $93.89\pm0.1632$ &  $0.9678\pm0.0128$ &  $0.3924\pm0.1138$ \\
    \multicolumn{1}{l|}{}	& OMADA-SE &  $0.0273\pm0.0012$ &  $\mathbf{94.988\pm0.1551}$ &  $0.9741\pm0.0112$ &  $0.5552\pm0.0428$ \\
   \multicolumn{1}{l|}{}	 & OMADA-SE-H &  $0.0281\pm0.001$ &  $94.258\pm0.1291$ &  $0.9119\pm0.0243$ &  $0.7551\pm0.022$ \\
    \multicolumn{1}{l|}{}	& OMADA-SE-U &  $0.0302\pm0.0014$ &  $94.618\pm0.0807$ &  $\mathbf{0.9786\pm0.0106}$ &  $\mathbf{0.336\pm0.0542}$ \\
    \midrule
\multicolumn{1}{l|}{WRN}  & Base &  $0.0341\pm0.0014$ &  $91.596\pm0.177$ &  $0.9022\pm0.0097$ &  $0.7221\pm0.0199$ \\
\multicolumn{1}{l|}{}    & OMADA &  $0.0208\pm0.0011$ &  $95.772\pm0.0421$ &  $0.9243\pm0.0411$ &  $0.7442\pm0.0315$ \\
\multicolumn{1}{l|}{}    & OMADA-H &  $\mathbf{0.0172\pm0.0004}$ &  $95.15\pm0.0908$ &  $0.921\pm0.0221$ &  $0.7155\pm0.0405$ \\
 \multicolumn{1}{l|}{}   & OMADA-U &  $0.0274\pm0.0011$ &  $95.446\pm0.1021$ &  $0.975\pm0.0053$ &  $\mathbf{0.4337\pm0.0755}$ \\
  \multicolumn{1}{l|}{}  & OMADA-SE &  $0.0207\pm0.0007$ &  $\mathbf{96.022\pm0.0928}$ &  $\mathbf{0.9833\pm0.0061}$ &  $0.5032\pm0.0364$ \\
 \multicolumn{1}{l|}{}   & OMADA-SE-H &  $0.0222\pm0.0004$ &  $95.882\pm0.0342$ &  $0.9643\pm0.0224$ &  $0.5907\pm0.0784$ \\
 \multicolumn{1}{l|}{}   & OMADA-SE-U &  $0.0231\pm0.0012$ &  $95.248\pm0.1169$ &  $0.9346\pm0.0137$ &  $0.7391\pm0.0196$ \\
\bottomrule
\end{tabular}
	\caption{We report the mean and standard deviations across $5$ runs for all OMADA variants. The means were also shown in Table~\ref{table1_ablations} and show the ACE, ACC, AUC and OOD-MMC of each network. DN denotes DenseNet. Additionally, this table shows the OOD-MMC for all the networks.}
	\label{appendix_table_ablations}
\end{table*}

\begin{figure*} [!ht]
	\centering
	\begin{subfigure}{.48\textwidth}
		\centering
		\includegraphics[width=0.95\textwidth, clip]{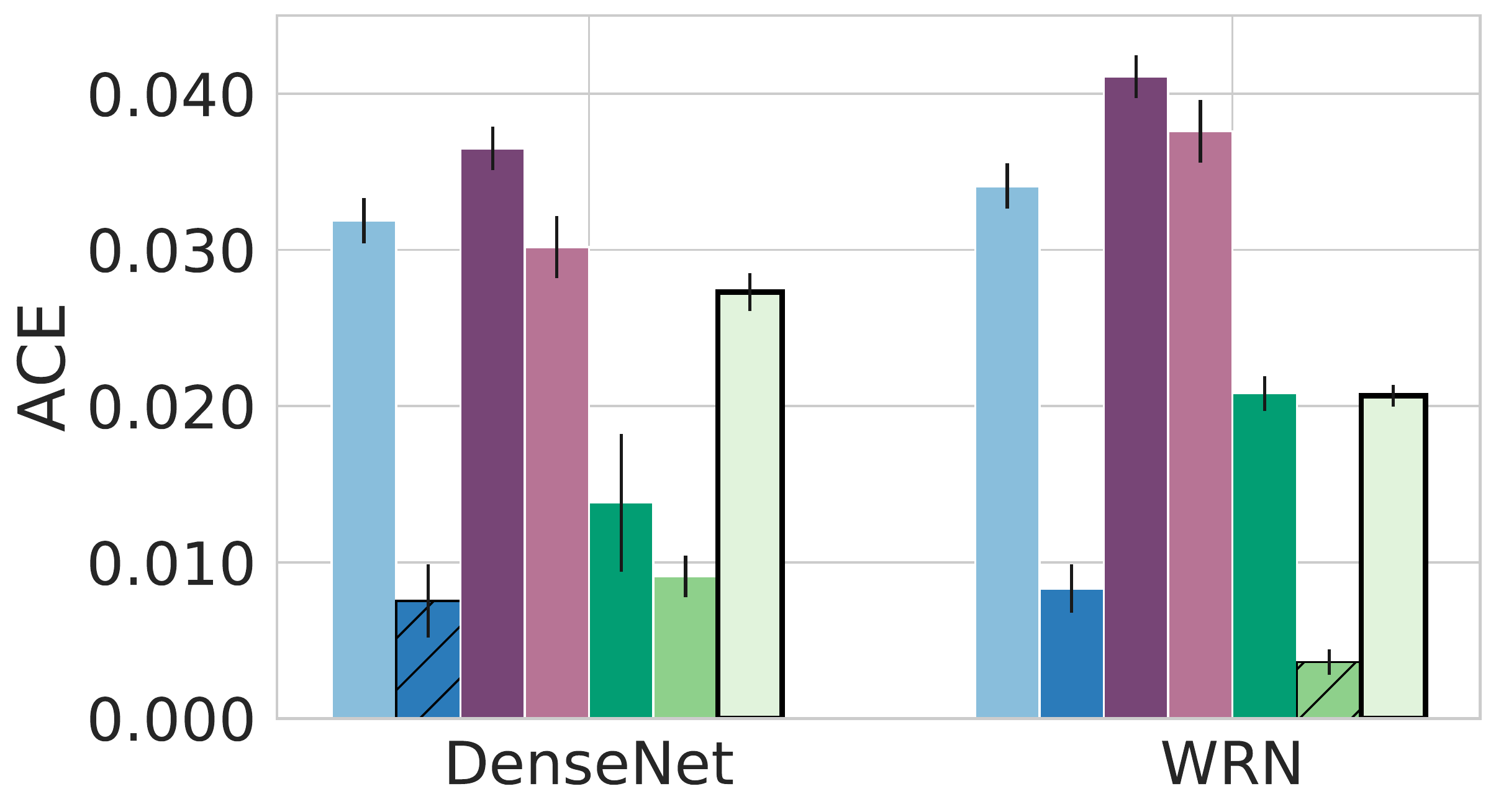}  
		\caption{CIFAR-10 ACE}
		\label{Afig4_stochastic_methods_ace}
	\end{subfigure}
	\begin{subfigure}{.48\textwidth}
		\centering
		\includegraphics[width=0.95\textwidth, clip]{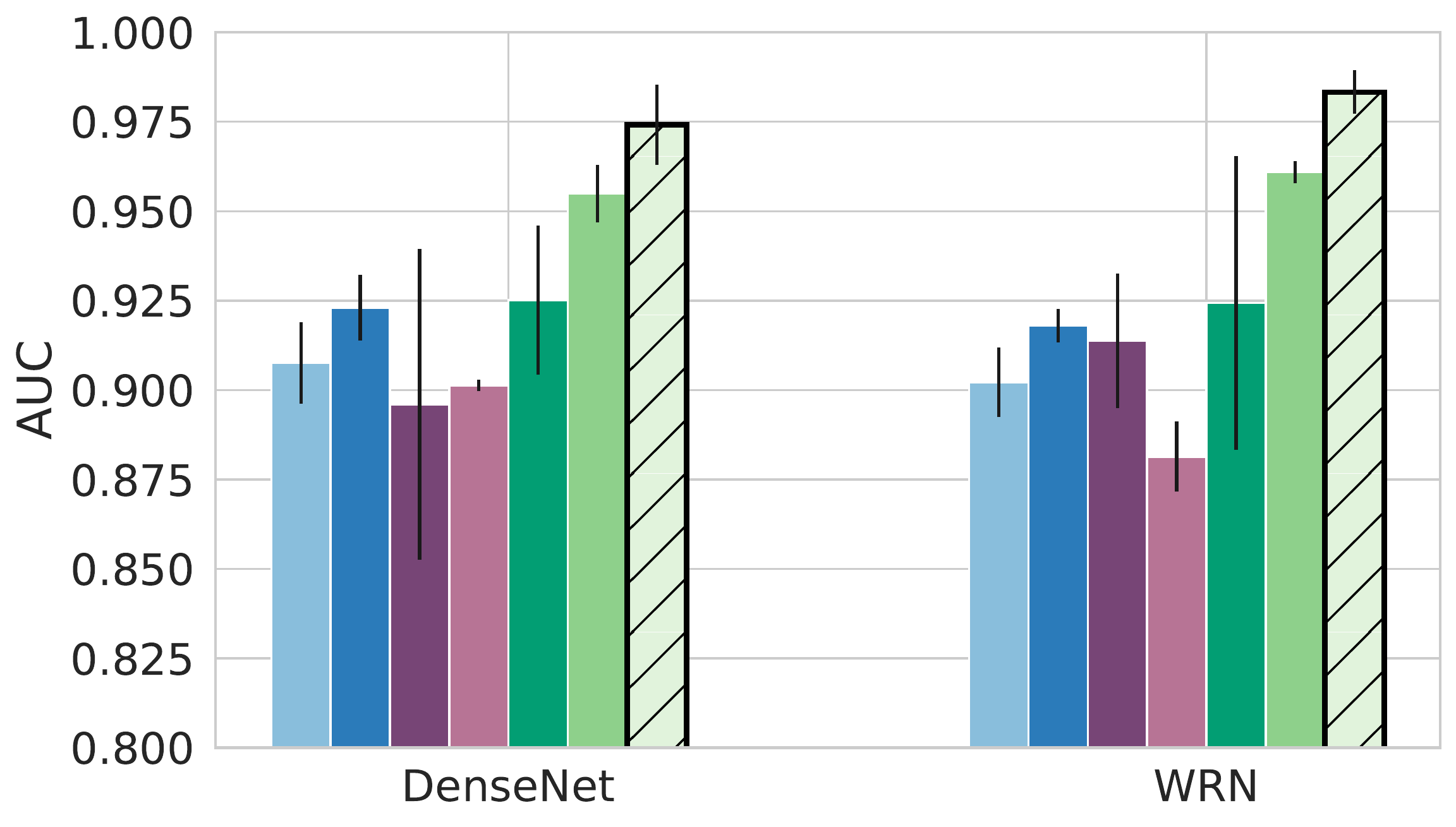}
		\caption{CIFAR-10 AUC}
		\label{Afig4_stochastic_methods_auc}
	\end{subfigure}
		\begin{subfigure}{.99\textwidth}
		\centering
		\includegraphics[width=0.95\textwidth, clip]{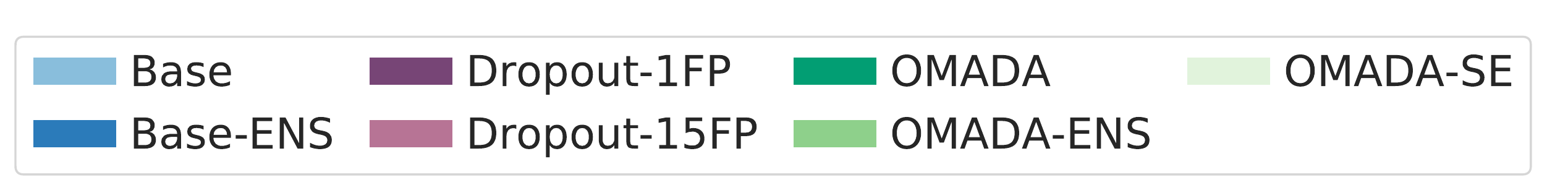}
		\label{legend}
	\end{subfigure}
	\caption{The figure depicts the ACE (a) and AUC (b) for CIFAR-10 on DenseNet and WRN. We denote the methods which involve an ensemble with ``*-ENS" (Ensemble of $5$ networks) and report the mean and standard deviation across $5$ sets of ensembles. The Dropout networks are specially trained networks with dropout and we report the results when using $1$ deterministic forward pass (Dropout-1FP) and $15$ stochastic forward passes (Dropout-15FP). Error bars are $\pm$ 1 std. dev. over 5 runs.	} 
	\label{AfigSTOCHASTIC}
\end{figure*}

\section{Stochastic Bayesian Neural Network Approximations} \label{stochastic_approximations}
Here we report the results of two stochastic Bayesian Neural Network Approximations: Ensembles and MC-Dropout (15 forward passes).
As these are orthogonal methods and can be applied to all methods, we compare our results with the Base network when applying these two stochastic approaches for uncertainty estimation.

Fig.~\ref{AfigSTOCHASTIC} shows the results for an Ensemble and MC-Dropout for CIFAR-10 on DenseNet and WRN.
Each ensemble entry reports the mean and standard deviation across $5$ ensembles, where each ensemble contains $5$ networks.
As none of the networks reported in the paper are trained with Dropout, we specially train DenseNet and WRN with dropout ($0.20$ and $0.30$, respectively) in order to compare against MC-Dropout.
As these networks can be considered to have a different network architecture compared to their no-dropout counter-parts, we report the ACE and AUC for a single deterministic forward pass through the network (Dropout-1FP) and compare this to the $15$ stochastic forward passes (Dropout-15FP).
We observe that for ACE on DenseNet, Base-ENS performs best, though after taking an ensemble of OMADA networks, we achieve similar performance.
However, for WRN, OMADA-ENS significantly surpasses Base-ENS.
This shows that ensembles help to improve network calibration, though come at the cost of expensive compute times during inference.
For ACE on both networks, MC-Dropout does not perform competitively.

On the other hand, when comparing the AUC numbers, we see that Base-ENS only slightly improves on Base and falls short of all OMADA-trained networks (with and without an ensemble on top). 
OMADA-ENS improves on OMADA alone, though interestingly it does not perform as well as OMADA-SE (which has much more samples with high entropy soft labels and confusing samples).

\section{When Does Temperature Scaling Help?} \label{temp_scaling_appendix}
\begin{figure*} [htp]
	\centering
	\begin{subfigure}{.48\textwidth}
		\centering
		\includegraphics[width=0.95\textwidth, clip]{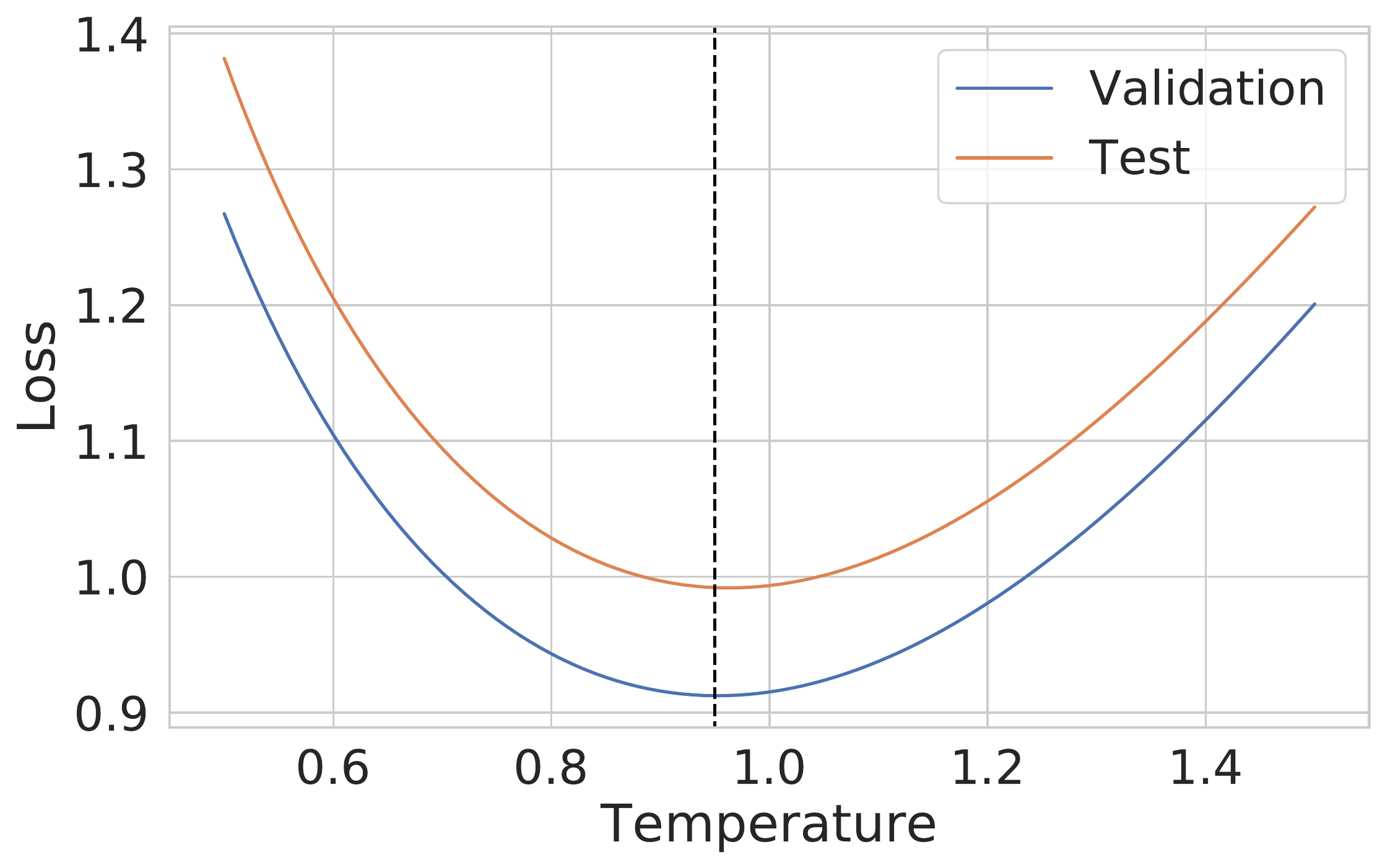}  
		\caption{CIFAR-100 + WRN Negative Log-Likelihood (NLL)}
		\label{Afig5_temp_scaling_loss}
	\end{subfigure}
	\begin{subfigure}{.48\textwidth}
		\centering
		\includegraphics[width=0.95\textwidth, clip]{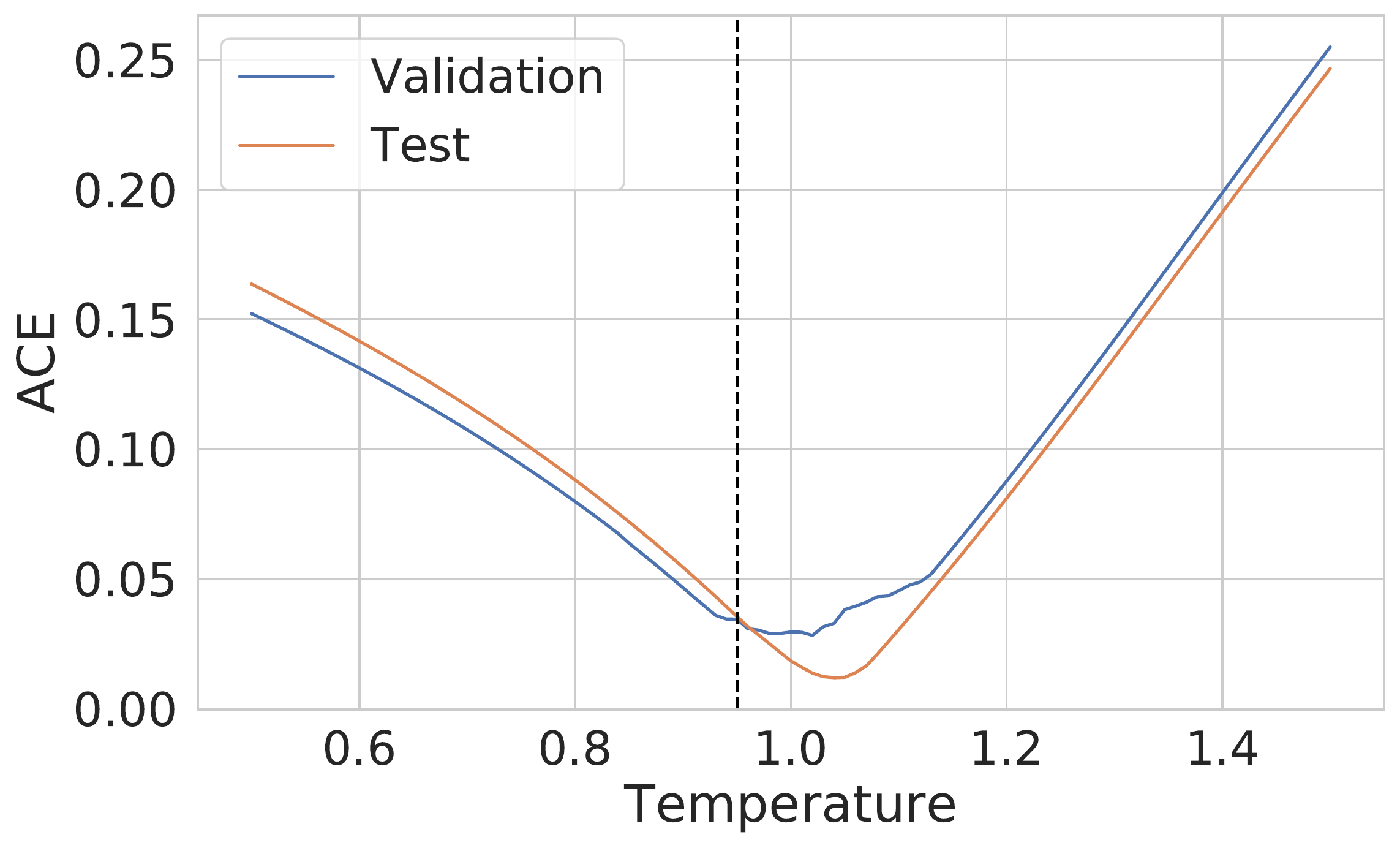}
		\caption{CIFAR-100 + WRN ACE}
		\label{Afig5_temp_scaling_ace}
	\end{subfigure}
	\caption{The figure depicts the NLL (a) and ACE (b) values when performing a grid-search for finding the best temperature T on the validation and test sets for CIFAR-100 on WRN. The vertical dashed black line shows the chosen temperature (T = 0.952) based on the lowest NLL on the validation set (i.e the optimized temperature). It can be seen that for both validation and test sets, the optimized temperatures do not minimize the ACE. In this case, in can be seen that not applying temperature scaling (T=1) would give a lower ACE.} 
	\label{AfigTEMPSCALE}
\end{figure*}

Temperature scaling is a simple method for improving network calibration. 
Interestingly, we observe that temperature scaling does not always improve performance; for networks which are fairly well calibrated already, the ACE gets worse by applying temperature scaling.
This suggests that the negative log likelihood (NLL) optimized by temperature scaling does not always correlate with a lower ACE (or ECE).
We show this phenomenon for WRN on CIFAR-100, where the optimized temperature increased the calibration error (ACE).
Fig.~\ref{AfigTEMPSCALE} shows the Negative Log-Likelihood (NLL) and ACE when performing a grid-search across temperatures. 
It can be seen that the best temperature (T = 0.952) based on the validation NLL (vertical dashed black line) does not minimize the ACE on the test nor the validation set. 
This shows that the NLL and ACE are not perfectly correlated (similar observations were made for ECE), and that a grid-search for the temperature based on the ACE might be an alternative option to find better temperatures.

\end{document}